\documentclass[sigconf]{acmart}
\AtBeginDocument{%
  }

\setcopyright{acmlicensed}
\copyrightyear{2018}
\acmYear{2018}
\acmDOI{XXXXXXX.XXXXXXX}
\acmConference[arXiv]{}
\acmISBN{}

\acmSubmissionID{1660}
\renewcommand\footnotetextcopyrightpermission[1]{}

\usepackage{multirow}
\usepackage{booktabs}  
\usepackage{amsmath}
\usepackage{subcaption}
\usepackage{bm}
\usepackage{float}
\usepackage{graphicx}
\usepackage{enumitem}
\usepackage{tipa}
\usepackage[font=small, skip=5pt]{caption}
\newcommand{\ours}{DynMask}
\newcommand{\msd}{MSD}

\usepackage[table]{xcolor}
\definecolor{bestblue}{RGB}{221,235,255}
\definecolor{secondblue}{RGB}{238,246,255}
\definecolor{groupblue}{RGB}{245,249,255}

\newcommand{\best}[1]{\cellcolor{bestblue}\textbf{#1}}
\newcommand{\second}[1]{\cellcolor{secondblue}#1}
\newcommand{\grouprow}[1]{\rowcolor{groupblue}\multicolumn{8}{l}{\textbf{#1}}}

\usepackage{algorithm}
\usepackage{algpseudocode}

\begin{document}

\title{Not All Frames Are Equal: Complexity-Aware Masked Motion Generation via Motion Spectral Descriptors}
\author{Pengfei Zhou}
\authornotemark[1]
\affiliation{%
\institution{Beihang University}
\city{Beijing}
\country{China}
}

\author{Xiangyue Zhang}
\authornote{Both authors contributed equally to this research.}
\orcid{0009-0002-5642-9474}
\affiliation{%
\institution{Wuhan University}
\city{Wuhan}
\country{China}
}

\author{Xukun Shen}
\affiliation{%
\institution{Beihang University}
\city{Beijing}
\country{China}
}

\author{Yong Hu}
\authornote{Corresponding Author.}
\affiliation{%
\institution{Beihang University}
\city{Beijing}
\country{China}
}
\renewcommand{\shortauthors}{Pengfei Zhou et al.}
\begin{abstract}
Masked generative models have become a strong paradigm for text-to-motion synthesis, but they still treat motion frames too uniformly during masking, attention, and decoding.
This is a poor match for motion, where local dynamic complexity varies sharply over time.
We show that current masked motion generators degrade disproportionately on dynamically complex motions, and that frame-wise generation error is strongly correlated with motion dynamics.
Motivated by this mismatch, we introduce the Motion Spectral Descriptor (MSD), a simple and parameter-free measure of local dynamic complexity computed from the short-time spectrum of motion velocity.
Unlike learned difficulty predictors, MSD is deterministic, interpretable, and derived directly from the motion signal itself. We use MSD to make masked motion generation complexity-aware.
In particular, MSD guides content-focused masking during training, provides a spectral similarity prior for self-attention, and can additionally modulate token-level sampling during iterative decoding.
Built on top of masked motion generators, our method, DynMask, improves motion generation most clearly on dynamically complex motions while also yielding stronger overall FID on HumanML3D and KIT-ML.
These results suggest that respecting local motion complexity is a useful design principle for masked motion generation. Project page: \url{https://xiangyue-zhang.github.io/DynMask}
\end{abstract}




\maketitle\section{Introduction}
\label{sec:intro}

\begin{figure}
  \includegraphics[width=0.45\textwidth]{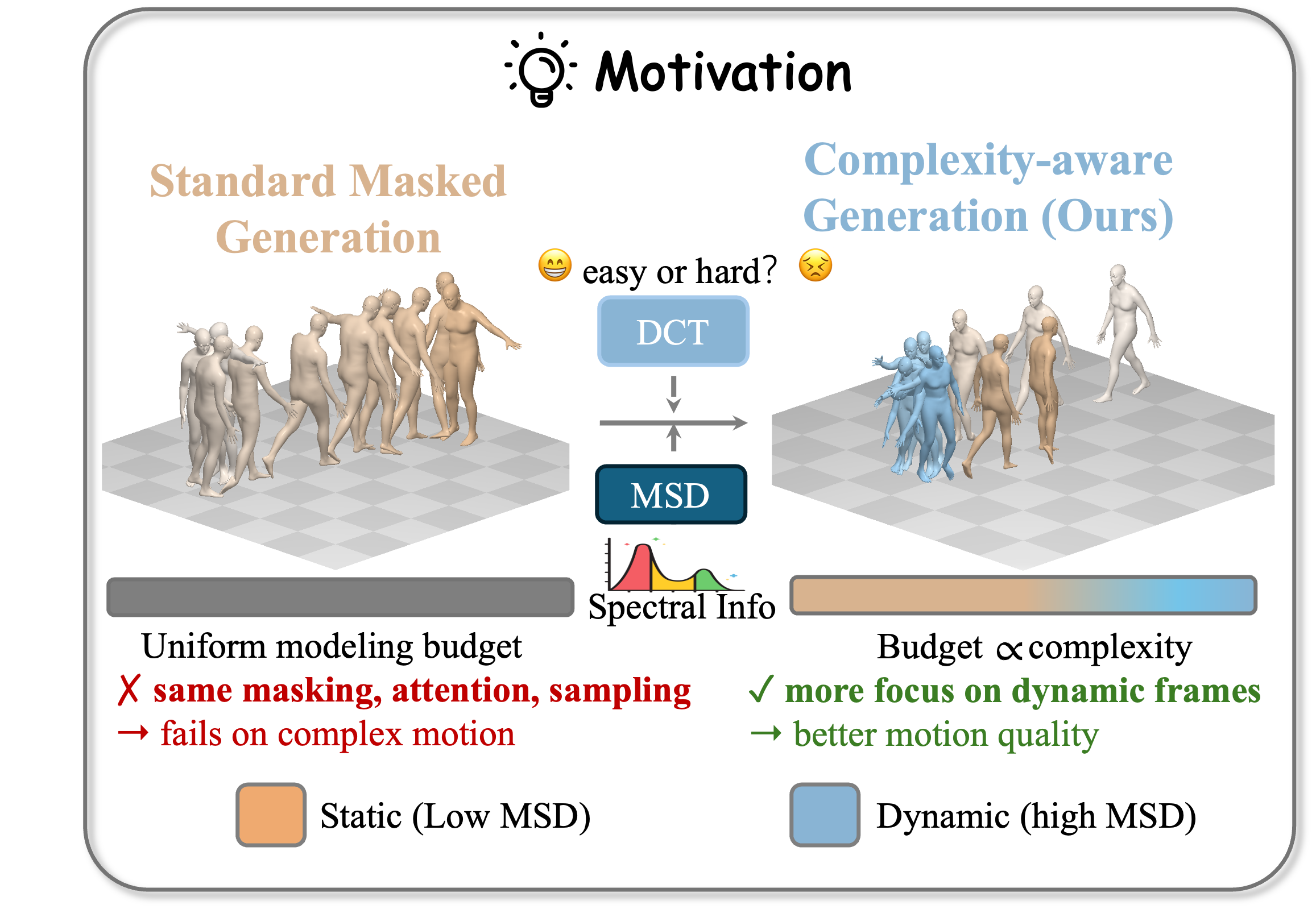}
  \caption{\textbf{Motivation.}
\emph{Left}: standard masked motion generation allocates uniform modeling budget across all frames, applying the same masking, attention, and sampling regardless of local motion difficulty---this leads to degraded quality on dynamically complex motion.
\emph{Right}: \ours{} uses the Motion Spectral Descriptor (\msd{}), a DCT-based per-frame complexity signal, to make masked generation complexity-aware: more supervision, stronger attention exchange, and broader decoding exploration are directed toward dynamically difficult frames, yielding better motion quality.}
  \label{fig:teaser}
\vspace{-15pt}
\end{figure}
Text-to-motion generation aims to synthesize realistic 3D human motion from natural language descriptions, with applications in animation, gaming, virtual reality, and embodied AI~\cite{tevet2022human,chen2023executing,zhang2023generating,guo2024momask,pinyoanuntapong2024mmm,pinyoanuntapong2024bamm}.
Among the major paradigms---diffusion models, autoregressive token models, and masked generative transformers---masked motion generation has emerged as an especially attractive choice.
It supports parallel decoding, is naturally compatible with editing and inpainting, and achieves strong benchmark performance by reconstructing motion tokens from partial context~\cite{guo2024momask,chang2022maskgit}.
However, a fundamental limitation remains: current masked motion generators treat frames too uniformly during mask allocation, attention, and iterative decoding.

This uniform treatment is a poor match for motion, whose local difficulty varies sharply over time.
Frames in static poses or smooth transitions are easy to infer from context, whereas frames at kicks, turns, jumps, or rapid phase transitions are much harder.
Yet standard masked motion generation applies nearly the same reconstruction policy to all of them~\cite{guo2024momask,yuan2024mogents}, causing errors to concentrate on the most dynamic parts of a sequence.

We refer to this mismatch as the \emph{dynamic uniformity assumption}: frames with sharply different local motion difficulty are processed with roughly uniform modeling budget.
This assumption pervades the entire masked pipeline---in training (mask positions ignore local difficulty), in attention (no structural bias toward dynamically compatible frames), and in decoding (uniform sampling regardless of uncertainty).
These are different manifestations of the same mismatch between non-uniform motion dynamics and uniform masked generation.

To address this issue, we seek a reusable signal that tells the model both \emph{where} motion is dynamically difficult and \emph{which} frames share similar local dynamics.
Such a signal should be motion-grounded, deterministic, and usable across different stages of the masked pipeline without requiring auxiliary models or additional supervision.
Based on this view, we introduce the \textbf{Motion Spectral Descriptor} (\msd{}), a parameter-free frame-level descriptor computed from the short-time spectrum of motion velocity.
For each frame, MSD produces both a scalar complexity summary and a structured spectral profile.
The scalar summary ranks dynamically difficult regions for mask allocation and decoding, while the spectral profile enables pairwise comparison of local motion dynamics for attention computation.

We then build \textbf{\ours{}} around this signal with a two-level design.
Its \textbf{core model} uses MSD for \emph{content-focused selection}, which allocates more reconstruction supervision to dynamically difficult frames, and for \emph{motion-aware attention}, which biases temporal information flow toward dynamically compatible regions.
Its \textbf{full model} further uses the same signal during iterative decoding through \emph{complexity-aware decoding}, where dynamically harder or more uncertain frames are decoded with broader exploration.
Rather than combining several unrelated heuristics, \ours{} uses one reusable motion-grounded signal to relax the dynamic uniformity assumption throughout masked motion generation.
Fig.~\ref{fig:teaser} illustrates the core motivation.

Our contributions are as follows:
\begin{itemize}[itemsep=0pt, parsep=0pt]
\sloppy
\item We identify the \emph{dynamic uniformity assumption} as a systematic limitation of masked motion generation: frames with sharply different local motion difficulty are treated too uniformly across training, attention, and decoding.
\item We propose the \textbf{Motion Spectral Descriptor} (\msd{}), a parameter-free signal that provides both scalar complexity ranking and structured spectral comparison, serving as a unified complexity measure throughout the masked pipeline.
\item We introduce \textbf{\ours{}}, whose core model uses MSD for content-focused mask selection and motion-aware attention, and whose full model further applies the same signal to complexity-aware decoding, achieving strong results particularly on dynamically complex motion.
\end{itemize}
\section{Related Work}
\label{sec:related}

\subsection{Motion Generation}
\label{sec:rw_motion}

Text-to-motion generation has been studied through several major paradigms.
Diffusion-based methods~\cite{tevet2022human,zhang2024motiondiffuse,chen2023executing,petrovich2022temos,zhou2024emdm,zhang2023remodiffuse,shafir2023priormdm,zhang2025energymogen} generate motion by iterative denoising under text conditioning.
Autoregressive methods~\cite{zhang2023generating,jiang2023motiongpt,zhang2024motionmamba,zhao2025dartcontrol} discretize motion into tokens and generate them sequentially.
More recently, masked motion generation~\cite{guo2024momask,pinyoanuntapong2024mmm,pinyoanuntapong2024bamm,zhang2024kmm} has emerged as a strong alternative by combining discrete tokenization with iterative parallel refinement, offering a favorable trade-off among generation quality, decoding efficiency, and editing flexibility.

Beyond text conditioning, motion generation has also been studied under richer control settings.
Inter-person conditioning~\cite{ma2025intersyn,ruizponce2024in2in,ruizponce2025mixermdm,wang2025timotion}, key-pose and trajectory guidance~\cite{karunratanakul2023guided,petrovich2024stmc,li2025frankenmotion,zhang2025robust,zhang2025towards}, fine-grained editing~\cite{zhang2024finemogen}, human--object interaction~\cite{wu2025hoi,xu2023interdiff,xu2024interdreamer,xu2025interact,xu2025intermimic,zeng2025chainhoi,zhang2024hoim3,zhang2022couch}, and scene-aware motion generation~\cite{pan2025tokenhsi,wang2025hsigpt,wang2024movesay} all study how to impose stronger structural or physical constraints on motion synthesis.
Related control settings also appear in audio-driven co-speech gesture generation~\cite{yi2023generating,liu2022beat,liu2024emage,ao2023gesturediffuclip,zhang2025echomask,zhang2025semtalk,zhang2026mitigating,zhi2023livelyspeaker,yang2026tokendancetokentotokenmusictodancegeneration} and music-conditioned dance generation~\cite{li2024lodge,li2023finedance,yang2025megadance,yang2025flowerdance,yang2025matchdance,yang2025mace,yang2024cohedancers,yang2024codancers,yang2024beatdance}.
These works differ in conditioning signals and generation objectives, but many share common backbones such as diffusion models or VQ-VAE-based discrete tokenization.

Our work builds on the masked text-to-motion paradigm.
Rather than introducing a new tokenizer or backbone, we focus on a limitation shared by current masked motion generators: they still treat motion frames too uniformly even though local dynamic complexity varies substantially over time.

\subsection{Masked Motion Modeling and Masking Strategies}
\label{sec:rw_masked}

Masked motion generation builds on masked prediction in language and vision~\cite{devlin2019bert,chang2022maskgit}.
A standard pipeline first tokenizes motion with VQ-VAE~\cite{ao2022rhythmic}, then reconstructs masked tokens conditioned on text and visible context, and finally performs iterative refinement at inference time.
Representative motion models include MoMask~\cite{guo2024momask}, MMM~\cite{pinyoanuntapong2024mmm}, BAMM~\cite{pinyoanuntapong2024bamm}, and KMM~\cite{zhang2024kmm}.

More broadly, masking strategy strongly affects what masked models learn.
Existing designs include random masking~\cite{devlin2019bert,dosovitskiy2020image,bao2021beit,chang2022maskgit,chang2023muse}, loss-based masking~\cite{bandara2023adamae,wang2023hard}, and attention-based masking~\cite{liu2023good,kakogeorgiou2022hide,hou2022milan}.
While these strategies show the value of non-uniform masking, they are usually semantic, task-dependent, or model-dependent.
For motion generation, a key difficulty comes from \emph{dynamic complexity}: frames can differ sharply in reconstruction difficulty because their local temporal dynamics are very different.
Yet current masked motion generators still remain largely complexity-agnostic, with generic mask schedules, content-only attention, and nearly uniform decoding policies.
Recent works such as HGM3~\cite{jeonghgm3} and EchoMask~\cite{zhang2025echomask} begin to question uniform masking.
Our work differs by introducing a motion-grounded complexity signal computed directly from the short-time spectrum of motion velocity, and by using this signal not only for mask allocation but also for motion-aware attention and, in the full model, decoding-time modulation.

\section{Method}
\label{sec:method}

\begin{figure*}[t]
\centering
\includegraphics[width=0.86\textwidth]{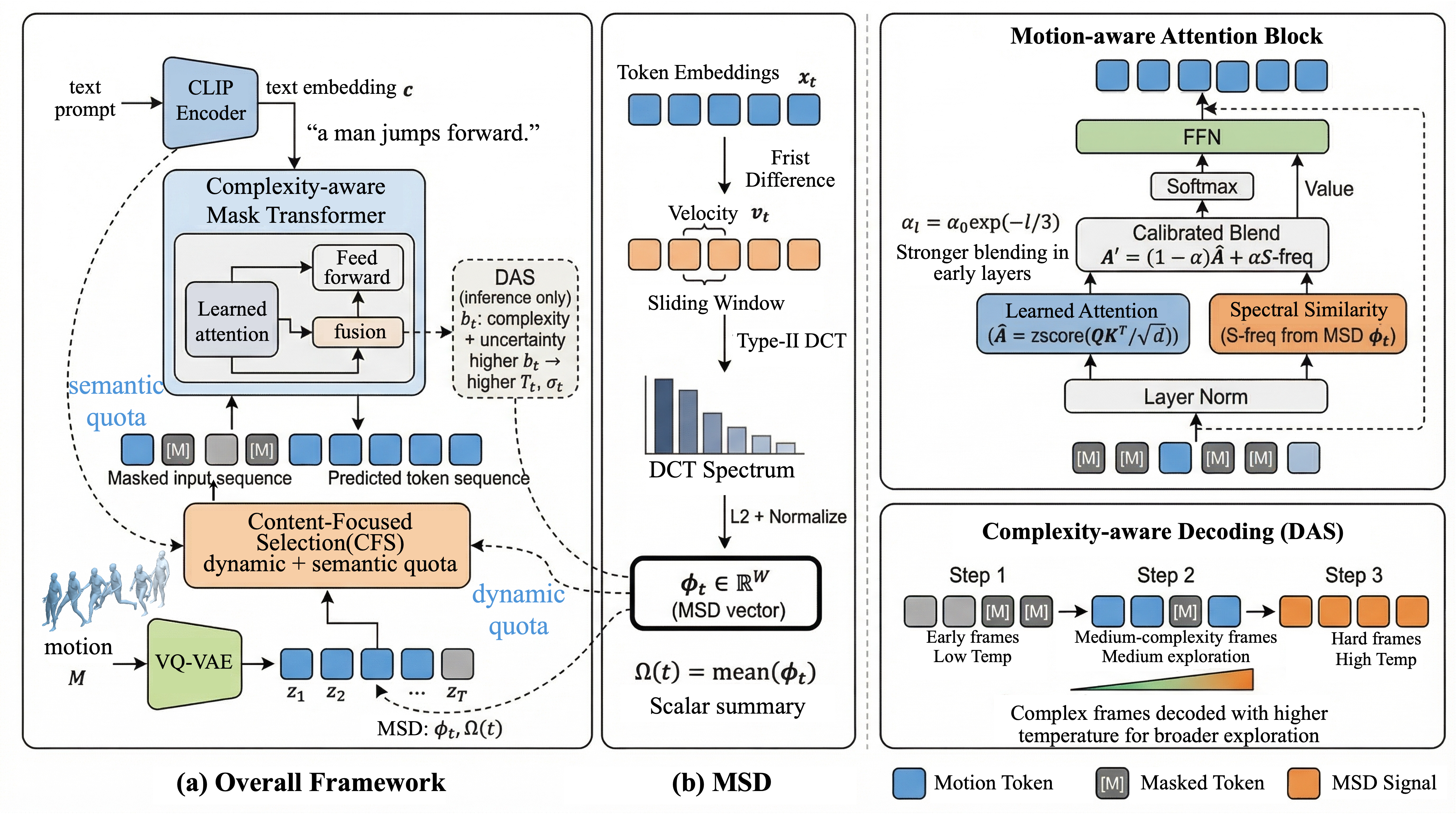}
\caption{\textbf{Overview of \ours{}.}
\textbf{(a)}~Core and full framework.
Given motion tokens from a VQ-VAE and a text condition from a frozen CLIP encoder, \ours{} computes one motion-grounded complexity signal, the Motion Spectral Descriptor (MSD), and reuses it throughout masked generation.
In the \textbf{core model}, MSD guides content-focused mask selection and motion-aware attention inside the masked transformer.
In the \textbf{full model}, the same signal is further used by complexity-aware decoding at inference time.
\textbf{(b)}~MSD computation.
For each frame, we compute token-embedding velocity, apply a sliding-window Type-II DCT, and obtain both a frame-level spectral descriptor $\boldsymbol{\phi}_t$ and a scalar complexity summary $\Omega(t)$.
\textbf{(c)}~Component details.
Motion-aware attention blends learned attention logits with MSD-derived spectral similarity using a layer-decayed coefficient, while complexity-aware decoding assigns higher temperature and noise to dynamically harder frames.}
\label{fig:method}
\end{figure*}

We build \ours{} around one motion-grounded signal, the Motion Spectral Descriptor (\msd{}), and use it to relax the dynamic uniformity assumption in masked motion generation.
The \textbf{core model} uses MSD in two places: complexity-aware mask selection and motion-aware attention.
The \textbf{full model} further uses the same signal during iterative decoding.
Fig.~\ref{fig:method} gives an overview.

\subsection{Preliminaries: Masked Motion Generation}
\label{sec:prelim}

We briefly review the masked motion generation pipeline that serves as the basis of our method~\cite{guo2024momask,chang2022maskgit}.

\noindent\textbf{Motion tokenization.}
Given a motion sequence $\mathbf{M} \in \mathbb{R}^{T \times D_m}$ with $T$ frames and motion feature dimension $D_m$, a VQ-VAE maps $\mathbf{M}$ to a discrete token sequence $\mathbf{z}=(z_1,\ldots,z_T)$, where each token $z_t \in \{1,\ldots,V\}$ indexes a codebook entry in $\mathcal{C}=\{\mathbf{e}_1,\ldots,\mathbf{e}_V\} \subset \mathbb{R}^{D}$.
The encoder $\mathcal{E}$ produces continuous latents $\hat{\mathbf{x}}_t=\mathcal{E}(\mathbf{M})_t$, which are quantized by nearest-neighbor lookup:
\begin{equation}
z_t = \operatorname*{argmin}_{v} \lVert \hat{\mathbf{x}}_t - \mathbf{e}_v \rVert_2 .
\end{equation}
The decoder $\mathcal{D}$ reconstructs motion from the quantized embeddings:
\begin{equation}
\hat{\mathbf{M}} = \mathcal{D}(\mathbf{e}_{z_1},\ldots,\mathbf{e}_{z_T}) .
\end{equation}

\noindent\textbf{Text-conditioned masked prediction.}
Following prior masked motion generation work~\cite{guo2024momask}, we encode the input text with a frozen CLIP text encoder and denote the resulting text condition as $\mathbf{c} \in \mathbb{R}^{E}$.
Given a token sequence $\mathbf{z}$, masked generation selects a subset of positions $\mathcal{S} \subset \{1,\ldots,T\}$ and replaces them with a special \texttt{[MASK]} symbol:
\[
\tilde{z}_t =
\begin{cases}
\texttt{[MASK]}, & t \in \mathcal{S}, \\
z_t, & \text{otherwise.}
\end{cases}
\]
A transformer $f_\theta$ takes the masked sequence $\tilde{\mathbf{z}}$ and text condition $\mathbf{c}$ as input and predicts the original tokens at masked positions.
Training minimizes the cross-entropy loss over the masked subset:
\begin{equation}
\mathcal{L}
=
-\sum_{t \in \mathcal{S}}
\log p_\theta(z_t \mid \tilde{\mathbf{z}}, \mathbf{c}) .
\label{eq:ce_loss}
\end{equation}
As in MaskGIT-style training~\cite{chang2022maskgit}, the total masking ratio follows a cosine schedule $\gamma(r)=\cos(\pi r/2)$ with $r \sim \mathcal{U}(0,1)$, while masked positions are usually sampled uniformly.

\noindent\textbf{Iterative decoding.}
At inference time, the token sequence is initialized as fully masked:
\[
\tilde{\mathbf{z}}^{(0)} = (\texttt{[MASK]},\ldots,\texttt{[MASK]}) .
\]
The model predicts masked tokens in parallel, keeps high-confidence predictions, and remasks uncertain positions over multiple refinement steps until the full sequence is decoded.
This pipeline is effective, but it still treats frames with very different local motion difficulty too uniformly.

\subsection{Motivation: Dynamic Complexity is Highly Non-Uniform}
\label{sec:principle}

The masked pipeline above is largely complexity-agnostic.
In practice, however, motion difficulty changes strongly over time: static poses and smooth transitions are easy to infer from context, whereas kicks, spins, jumps, and abrupt transitions are much harder.
Our empirical analysis (Sec.~\ref{sec:exp_complexity}) confirms that masked generators degrade much more on dynamic actions, with the largest failures concentrated around frames with strong local dynamics.
We refer to this issue as the \emph{dynamic uniformity assumption}, and our goal is to relax it by introducing a motion-grounded complexity signal to guide where and how the pipeline allocates its modeling effort.

\subsection{Motion Spectral Descriptor}
\label{sec:msd}

\noindent\textbf{Overview.}
We measure local dynamic complexity from the short-time spectrum of motion velocity.
The key intuition is simple: static or slowly changing motion has concentrated low-frequency energy, while fast and irregular motion produces richer temporal spectrum.
This makes short-time spectral content a natural descriptor of local dynamic complexity.

\noindent\textbf{Feature-space velocity.}
Let $\mathbf{x}_t=\mathrm{Embed}(z_t) \in \mathbb{R}^{D}$ denote the token embedding at frame $t$.
We first compute the temporal first difference:
\begin{equation}
\mathbf{v}_t = \mathbf{x}_t - \mathbf{x}_{t-1},
\qquad t=2,\ldots,T ,
\label{eq:velocity}
\end{equation}
with $\mathbf{v}_1=\mathbf{v}_2$ for boundary handling.
We use velocity rather than position because our target is dynamic complexity, namely how rapidly and irregularly motion changes over time.

\noindent\textbf{Sliding-window DCT.}
For each frame $t$, we extract a local temporal window of velocity vectors of length $W$, centered at $t$, and apply a Type-II DCT~\cite{ahmed1974discrete} along the temporal axis:
\begin{equation}
F_{k,d}
=
\sum_{n=0}^{W-1}
v_{n,d}
\cos\Bigl(
\frac{\pi}{W}\Bigl(n+\tfrac{1}{2}\Bigr)k
\Bigr),
\qquad k=0,\ldots,W-1.
\label{eq:dct}
\end{equation}
where $d$ indexes feature dimensions.
We then aggregate the spectrum across dimensions using the L2 norm:
\begin{equation}
f_k = \sqrt{\sum_d F_{k,d}^2} .
\end{equation}
The resulting frequency response is L2-normalized to form the \textbf{Motion Spectral Descriptor} (\msd{}) at frame $t$:
\begin{equation}
\boldsymbol{\phi}_t
=
\frac{(f_0,f_1,\ldots,f_{W-1})}
{\lVert (f_0,f_1,\ldots,f_{W-1}) \rVert_2 + \epsilon}
\in \mathbb{R}^{W} .
\label{eq:msd}
\end{equation}

\noindent\textbf{Scalar complexity summary.}
The full vector $\boldsymbol{\phi}_t$ captures local spectral shape and is useful for comparing pairs of frames.
For scalar complexity ranking, we use the mean spectral activation:
\begin{equation}
\Omega(t)
=
\bar{\phi}_t
=
\frac{1}{W}\sum_{k=0}^{W-1}\phi_{t,k} .
\label{eq:omega}
\end{equation}
We use $\Omega(t)$ only as a compact proxy for local dynamic complexity.
It is not meant to measure semantic importance or full motion difficulty, but specifically the dynamic complexity of local motion evolution.

\noindent\textbf{Why not use velocity alone.}
Joint velocity magnitude is a strong scalar proxy for motion difficulty, but it only measures how large the movement is.
In contrast, MSD provides both
(i) a scalar complexity summary $\Omega(t)$ for ranking difficult regions and
(ii) a structured spectral profile $\boldsymbol{\phi}_t$ for pairwise frame comparison.
This second property is important because our method uses the same signal not only for mask allocation, but also for motion-aware attention.

\subsection{Core Model: Complexity-Aware Mask Selection}
\label{sec:cfs}

As shown in Algorithms~\ref{alg:train}, Our first and most direct use of MSD is to guide where reconstruction supervision should be concentrated during training.
The default masked pipeline selects masked positions uniformly at random.
For motion, this underuses dynamically difficult regions, even though these are often the regions where reconstruction is hardest.

\noindent\textbf{Content-Focused Selection.}
We score each frame from two complementary views.
The first is dynamic complexity derived from MSD:
\begin{equation}
s_t^{\text{dyn}}
=
\text{z-score}(\bar{\phi}_t) ,
\label{eq:sdyn}
\end{equation}
where $\bar{\phi}_t$ is defined in Eq.~\eqref{eq:omega}.
The second is semantic salience with respect to the text condition:
\begin{equation}
s_t^{\text{sem}}
=
\text{z-score}
\left(
\frac{
\langle \mathbf{x}_t, \mathbf{c} \rangle
}{
\lVert \mathbf{x}_t \rVert \,
\lVert \mathbf{c} \rVert
}
\right) .
\label{eq:ssem}
\end{equation}
The dynamic score emphasizes frames with complex local motion, while the semantic score preserves frames that are strongly aligned with the text condition.

Given a total masking budget $K$, we allocate two quotas:
$(1-\lambda)K$ positions are selected from frames with high dynamic-complexity score, and $\lambda K$ positions are selected from frames with high semantic-salience score.
An optional temporal expansion radius $r_{\text{exp}}$ extends each selected position to nearby frames, which improves coverage around important temporal regions.

A key point is that we change \emph{which} positions are masked, but not \emph{how many}.
The total masking ratio still follows the standard cosine schedule.
This isolates the role of complexity-aware supervision allocation from the overall amount of masking.

\subsection{Core Model: Motion-Aware Attention}
\label{sec:attn}

Our second core use of MSD is to provide a motion-grounded bias for self-attention.
Standard attention measures similarity in a learned feature space.
For motion, however, two frames may play similar dynamic roles even if their raw embeddings are not close.
Examples include repeated gait phases, similar airborne phases, or temporally distant segments with comparable rhythmic structure.

\noindent\textbf{Spectral similarity.}
Given MSD vectors $\boldsymbol{\phi}_i$ and $\boldsymbol{\phi}_j$ for frames $i$ and $j$, we define their spectral similarity as
\begin{equation}
S^{\text{freq}}_{i,j}
=
-\frac{1}{\tau}\sum_{k=0}^{W-1} w_k \bigl(\phi_{i,k}-\phi_{j,k}\bigr)^2 ,
\label{eq:freq_sim}
\end{equation}
where $\tau > 0$ is a scale factor and
\begin{equation}
w_k
=
\frac{\exp(-k/3)}{\sum_{k'} \exp(-k'/3)}
\end{equation}
assigns larger weight to lower-frequency bands, which tend to be more stable indicators of local motion structure.

\noindent\textbf{Attention fusion.}
Let $A_{i,j}$ denote the standard self-attention logits.
We blend the learned attention logits with the spectral similarity prior:
\begin{equation}
A'_{i,j}
=
(1-\alpha_\ell)\hat{A}_{i,j}
+
\alpha_\ell \hat{S}^{\text{freq}}_{i,j} ,
\label{eq:blend}
\end{equation}
where $\hat{A}_{i,j}$ and $\hat{S}^{\text{freq}}_{i,j}$ are z-score normalized over valid positions, and $\alpha_\ell$ is a layer-dependent fusion weight:
\begin{equation}
\alpha_\ell = \alpha_0 \exp(-\ell/3) .
\label{eq:alpha}
\end{equation}
This schedule places stronger motion prior in earlier layers, where coarse temporal organization is formed, and gives later layers more freedom to refine motion from learned content.

\noindent\textbf{Motion-aware attention block.}
The fused logits $A'_{i,j}$ are passed through the usual attention computation to obtain the attention map, which is then applied to value features before the feed-forward layer.
In this way, the spectral prior changes \emph{how} information is routed across time rather than adding a separate supervision term.

\subsection{Full Model Extension: Complexity-Aware Decoding}
\label{sec:das}

 The two components above define the \textbf{core model} of \ours{}.
As illustrated in Algorithms~\ref{alg:infer}, in the \textbf{full model}, we additionally use MSD during iterative decoding.
This decoding-time component is optional.
It is not required for the main idea of complexity-aware supervision and motion-aware attention, but it provides an additional way to make generation sensitive to local motion difficulty at inference time.

\noindent\textbf{Exploration score.}
At decoding step $s$, we compute a frame-wise exploration score that combines dynamic complexity and model uncertainty:
\begin{equation}
b_t = \sigma\bigl(\lambda_D \hat{k}_t + (1-\lambda_D)\hat{u}_t\bigr).
\label{eq:blend_u}
\end{equation}
Here, $\sigma(\cdot)$ denotes the sigmoid function. The normalized complexity term is
\begin{equation}
\hat{k}_t = \operatorname{zscore}(\bar{\phi}_t).
\label{eq:khat}
\end{equation}
The normalized uncertainty term is
\begin{equation}
\hat{u}_t = \operatorname{zscore}(1 - \max_v p_{t,v}).
\label{eq:uhat}
\end{equation}

\noindent\textbf{Token-level temperature modulation.}
We use $b_t$ to modulate token-level sampling temperature and noise:
\begin{equation}
T_t = T^{\mathrm{global}} (1 + \beta b_t).
\label{eq:adapt_temp}
\end{equation}
\begin{equation}
\sigma_t = \sigma_{\max} b_t.
\label{eq:adapt_noise}
\end{equation}

\noindent\textbf{Token-level temperature modulation.}
We use $b_t$ to modulate token-level sampling temperature and noise:
\begin{align}
T_t &= T^{\text{global}} \cdot (1 + \beta b_t) ,
\label{eq:adapt_temp} \\
\sigma_t &= \sigma_{\max} \cdot b_t .
\label{eq:adapt_noise}
\end{align}
Here $T^{\text{global}}$ is a step-dependent global temperature, $\beta$ controls the strength of adaptation, and $\sigma_{\max}$ is the maximum noise scale.
Frames that are more complex or more uncertain are allowed broader exploration, while simpler and more confident frames are decoded more conservatively.

\begin{algorithm}[t]
\caption{\ours{} training (core model)}
\label{alg:train}
\small
\begin{algorithmic}[1]
\Require Token sequence $\mathbf{z}$, text condition $\mathbf{c}$
\State Compute token embeddings $\mathbf{x}_t \gets \mathrm{Embed}(z_t)$
\State Compute velocity $\mathbf{v}_t \gets \mathbf{x}_t - \mathbf{x}_{t-1}$
\State Compute \msd{} $\boldsymbol{\phi}_t$ using sliding-window DCT
\State Compute dynamic score $s_t^{\text{dyn}}$ and semantic score $s_t^{\text{sem}}$
\State Select $K$ mask positions using dual-quota Content-Focused Selection
\State Apply BERT-style masking to the selected positions
\State Forward the masked sequence with motion-aware attention
\State Compute cross-entropy loss on masked positions
\end{algorithmic}
\end{algorithm}

\begin{algorithm}[t]
\caption{\ours{} inference (full model)}
\label{alg:infer}
\small
\begin{algorithmic}[1]
\Require Text condition $\mathbf{c}$, target length $T$, decoding steps $S$
\State Initialize fully masked sequence $\hat{\mathbf{z}} \gets [\texttt{MASK}]^T$
\For{$s = 1, \ldots, S$}
    \State Compute or update \msd{} $\boldsymbol{\phi}_t$ from the current decoded sequence
    \State Forward pass with motion-aware attention to obtain logits $\mathbf{p}_t$
    \State Compute exploration score $b_t$ from complexity and uncertainty
    \State Set token-level temperature $T_t$ and noise $\sigma_t$
    \State Sample token predictions and keep the most confident positions
\EndFor
\State Decode tokens to motion using the VQ-VAE decoder
\end{algorithmic}
\end{algorithm}
\section{Experiments}
\label{sec:exp}

We evaluate \ours{} on HumanML3D and KIT-ML, comparing against representative methods from multiple generation paradigms.
We then conduct systematic ablations to isolate the contribution of each design choice, analyze where the improvements concentrate across motion complexity levels, and validate the results through qualitative comparison and a user study.

\subsection{Setup}
\label{sec:setup}

\noindent\textbf{Datasets.}
We evaluate on three standard motion datasets.
\textbf{HumanML3D}~\cite{guo2022generating} is our main benchmark and provides paired motion-text examples with the standard train/val/test split.
\textbf{KIT-ML}~\cite{plappert2016kit} serves as a second benchmark with different motion and language statistics.
We also use \textbf{BABEL}~\cite{punnakkal2021babel} for cross-dataset transfer analysis.

\noindent\textbf{Metrics.}
Following standard text-to-motion evaluation~\cite{guo2022generating}, we report R-Precision at top-1, top-2, and top-3, Fr\'echet Inception Distance (\textbf{FID}), Multimodal Distance (\textbf{MM-Dist}), Diversity, and MModality when available.
All metrics are computed using the standard motion and text feature extractors used in prior work.
We report averages over repeated evaluations, and confidence intervals are shown in the tables.

\noindent\textbf{Implementation details.}
The \textbf{core model} of \ours{} consists of complexity-aware mask selection and motion-aware attention.
The \textbf{full model} additionally includes complexity-aware decoding at inference time.
All main experiments use the same default hyperparameters unless a specific ablation states otherwise.

\subsection{Quantitative results}
\label{sec:exp_sota}

We compare \ours{} with representative methods from different generation paradigms on HumanML3D and KIT-ML.
Our main scope is masked motion generation, while other paradigms are included for broader context.

\begin{table*}[t]
\centering
\caption{\textbf{Comparison on HumanML3D.}
Methods are grouped by generation paradigm.
Our main scope is improvement within the masked motion generation family, while other paradigms are reported for broader context.
Blue background indicates the best and second-best generated results.}
\label{tab:sota_h3d}
\tiny
\footnotesize
\setlength{\tabcolsep}{4pt}
\begin{tabular}{lccccccc}
\toprule
Method & Top-1$\uparrow$ & Top-2$\uparrow$ & Top-3$\uparrow$ & FID$\downarrow$ & MM-Dist$\downarrow$ & Diversity$\uparrow$ & MModality$\uparrow$ \\
\midrule
\grouprow{Diffusion-based Methods} \\
MotionDiffuse~\cite{zhang2024motiondiffuse} & 0.491{\scriptsize$\pm$.001} & 0.681{\scriptsize$\pm$.001} & 0.782{\scriptsize$\pm$.001} & 0.630{\scriptsize$\pm$.001} & 3.113{\scriptsize$\pm$.001} & 9.410{\scriptsize$\pm$.049} & 1.553{\scriptsize$\pm$.042} \\
MLD~\cite{chen2023executing} & 0.481{\scriptsize$\pm$.003} & 0.673{\scriptsize$\pm$.003} & 0.772{\scriptsize$\pm$.002} & 0.473{\scriptsize$\pm$.013} & 3.196{\scriptsize$\pm$.010} & 9.724{\scriptsize$\pm$.082} & 2.413{\scriptsize$\pm$.079} \\
ReMoDiffuse~\cite{zhang2023remodiffuse} & 0.510{\scriptsize$\pm$.003} & 0.698{\scriptsize$\pm$.006} & 0.795{\scriptsize$\pm$.004} & 0.103{\scriptsize$\pm$.004} & 2.974{\scriptsize$\pm$.016} & 9.018{\scriptsize$\pm$.075} & --- \\
DiverseMotion~\cite{lou2023diversemotion} & 0.496{\scriptsize$\pm$.004} & 0.687{\scriptsize$\pm$.004} & 0.783{\scriptsize$\pm$.004} & 0.070{\scriptsize$\pm$.004} & 3.063{\scriptsize$\pm$.011} & 9.551{\scriptsize$\pm$.068} & --- \\
\midrule
\grouprow{Other Token-based Methods} \\
Fg-T2M~\cite{wang2023fg} & 0.492{\scriptsize$\pm$.002} & 0.683{\scriptsize$\pm$.003} & 0.783{\scriptsize$\pm$.002} & 0.243{\scriptsize$\pm$.019} & 3.109{\scriptsize$\pm$.007} & 9.278{\scriptsize$\pm$.072} & 1.614{\scriptsize$\pm$.049} \\
AttT2M~\cite{kakogeorgiou2022hide} & 0.499{\scriptsize$\pm$.003} & 0.690{\scriptsize$\pm$.002} & 0.786{\scriptsize$\pm$.002} & 0.112{\scriptsize$\pm$.006} & 3.038{\scriptsize$\pm$.007} & 9.700{\scriptsize$\pm$.090} & 2.452{\scriptsize$\pm$.051} \\
LaMP~\cite{li2024lamp} & 0.557{\scriptsize$\pm$.003} & 0.751{\scriptsize$\pm$.002} & 0.843{\scriptsize$\pm$.001} & 0.032{\scriptsize$\pm$.002} & 2.759{\scriptsize$\pm$.007} & 9.571{\scriptsize$\pm$.069} & --- \\
\midrule
\grouprow{Masked Motion Generation} \\
MMM~\cite{pinyoanuntapong2024mmm} & 0.515{\scriptsize$\pm$.002} & 0.708{\scriptsize$\pm$.002} & 0.804{\scriptsize$\pm$.002} & 0.089{\scriptsize$\pm$.005} & 2.926{\scriptsize$\pm$.007} & 9.577{\scriptsize$\pm$.050} & 1.226{\scriptsize$\pm$.035} \\
MoMask~\cite{guo2024momask} & 0.521{\scriptsize$\pm$.003} & 0.713{\scriptsize$\pm$.003} & 0.807{\scriptsize$\pm$.003} & 0.045{\scriptsize$\pm$.003} & 2.958{\scriptsize$\pm$.008} & --- & 1.241{\scriptsize$\pm$.040} \\
BAMM~\cite{pinyoanuntapong2024bamm} & 0.525{\scriptsize$\pm$.003} & 0.720{\scriptsize$\pm$.003} & 0.813{\scriptsize$\pm$.003} & 0.055{\scriptsize$\pm$.003} & 2.919{\scriptsize$\pm$.008} & \second{9.717{\scriptsize$\pm$.089}} & \best{1.687{\scriptsize$\pm$.051}} \\
ParCo~\cite{zou2024parco} & 0.515{\scriptsize$\pm$.003} & 0.706{\scriptsize$\pm$.003} & 0.801{\scriptsize$\pm$.002} & 0.109{\scriptsize$\pm$.005} & 2.927{\scriptsize$\pm$.008} & 9.576{\scriptsize$\pm$.088} & --- \\
HGM3~\cite{jeonghgm3} & \best{0.535{\scriptsize$\pm$.002}} & \best{0.726{\scriptsize$\pm$.002}} & \best{0.822{\scriptsize$\pm$.002}} & \second{0.036{\scriptsize$\pm$.002}} & \second{2.904{\scriptsize$\pm$.008}} & 9.545{\scriptsize$\pm$.091} & 1.206{\scriptsize$\pm$.051} \\
\midrule
\textbf{\ours{} (ours)} & \second{0.528{\scriptsize$\pm$.002}} & \second{0.721{\scriptsize$\pm$.002}} & \second{0.814{\scriptsize$\pm$.002}} & \best{0.028{\scriptsize$\pm$.005}} & \best{2.879{\scriptsize$\pm$.006}} & \best{9.73{\scriptsize$\pm$.040}} & \second{1.311{\scriptsize$\pm$.018}} \\
\bottomrule
\end{tabular}
\vspace{-15pt}
\end{table*}

\begin{table*}[t]
\centering
\caption{\textbf{Comparison on KIT-ML.}
Methods are grouped by generation paradigm.
Our main scope is masked motion generation, and other paradigms are included for reference.
Blue background indicates the best and second-best generated results.}
\label{tab:sota_kit}
\tiny
\footnotesize
\setlength{\tabcolsep}{4pt}
\begin{tabular}{lccccccc}

\toprule
Method & Top-1$\uparrow$ & Top-2$\uparrow$ & Top-3$\uparrow$ & FID$\downarrow$ & MM-Dist$\downarrow$ & Diversity$\uparrow$ & MModality$\uparrow$ \\
\midrule
\grouprow{Diffusion-based Methods} \\
MotionDiffuse~\cite{zhang2024motiondiffuse} & 0.417{\scriptsize$\pm$.004} & 0.621{\scriptsize$\pm$.004} & 0.739{\scriptsize$\pm$.004} & 1.954{\scriptsize$\pm$.064} & 2.958{\scriptsize$\pm$.005} & 11.10{\scriptsize$\pm$.143} & 0.730{\scriptsize$\pm$.013} \\
MLD~\cite{chen2023executing} & 0.390{\scriptsize$\pm$.008} & 0.609{\scriptsize$\pm$.008} & 0.734{\scriptsize$\pm$.007} & 0.404{\scriptsize$\pm$.027} & 3.204{\scriptsize$\pm$.027} & 10.80{\scriptsize$\pm$.117} & 2.192{\scriptsize$\pm$.071} \\
ReMoDiffuse~\cite{zhang2023remodiffuse} & 0.427{\scriptsize$\pm$.014} & 0.641{\scriptsize$\pm$.004} & 0.765{\scriptsize$\pm$.005} & 0.155{\scriptsize$\pm$.006} & 2.814{\scriptsize$\pm$.012} & 10.80{\scriptsize$\pm$.105} & --- \\
DiverseMotion~\cite{lou2023diversemotion} & 0.416{\scriptsize$\pm$.005} & 0.637{\scriptsize$\pm$.008} & 0.760{\scriptsize$\pm$.011} & 0.468{\scriptsize$\pm$.098} & 2.892{\scriptsize$\pm$.041} & 10.873{\scriptsize$\pm$.101} & --- \\
\midrule
\grouprow{Other Token-based Methods} \\
Fg-T2M~\cite{wang2023fg} & 0.418{\scriptsize$\pm$.005} & 0.626{\scriptsize$\pm$.004} & 0.745{\scriptsize$\pm$.004} & 0.571{\scriptsize$\pm$.047} & 3.114{\scriptsize$\pm$.015} & 10.93{\scriptsize$\pm$.083} & 1.019{\scriptsize$\pm$.029} \\
AttT2M~\cite{kakogeorgiou2022hide} & 0.413{\scriptsize$\pm$.006} & 0.632{\scriptsize$\pm$.006} & 0.751{\scriptsize$\pm$.006} & 0.870{\scriptsize$\pm$.039} & 3.039{\scriptsize$\pm$.021} & 10.96{\scriptsize$\pm$.123} & 2.281{\scriptsize$\pm$.047} \\
LaMP~\cite{li2024lamp} & 0.479{\scriptsize$\pm$.006} & 0.691{\scriptsize$\pm$.005} & 0.826{\scriptsize$\pm$.005} & 0.141{\scriptsize$\pm$.013} & 2.704{\scriptsize$\pm$.018} & 10.929{\scriptsize$\pm$.101} & --- \\
\midrule
\grouprow{Masked Motion Generation} \\
MMM~\cite{pinyoanuntapong2024mmm} & 0.404{\scriptsize$\pm$.005} & 0.621{\scriptsize$\pm$.005} & 0.744{\scriptsize$\pm$.004} & 0.316{\scriptsize$\pm$.028} & 2.977{\scriptsize$\pm$.019} & 10.910{\scriptsize$\pm$.101} & 1.232{\scriptsize$\pm$.039} \\
MoMask~\cite{guo2024momask} & 0.433{\scriptsize$\pm$.007} & 0.656{\scriptsize$\pm$.005} & 0.781{\scriptsize$\pm$.005} & 0.204{\scriptsize$\pm$.011} & 2.779{\scriptsize$\pm$.022} & --- & 1.131{\scriptsize$\pm$.043} \\
BAMM~\cite{pinyoanuntapong2024bamm} & 0.438{\scriptsize$\pm$.009} & 0.661{\scriptsize$\pm$.009} & 0.788{\scriptsize$\pm$.005} & 0.183{\scriptsize$\pm$.013} & 2.723{\scriptsize$\pm$.026} & \best{11.008{\scriptsize$\pm$.094}} & \best{1.609{\scriptsize$\pm$.065}} \\
ParCo~\cite{zou2024parco} & 0.430{\scriptsize$\pm$.004} & 0.649{\scriptsize$\pm$.007} & 0.772{\scriptsize$\pm$.006} & 0.453{\scriptsize$\pm$.027} & 2.820{\scriptsize$\pm$.028} & \second{10.95{\scriptsize$\pm$.094}} & --- \\
HGM3~\cite{jeonghgm3} & \best{0.444{\scriptsize$\pm$.007}} & \second{0.664{\scriptsize$\pm$.007}} & \best{0.791{\scriptsize$\pm$.006}} & 0.176{\scriptsize$\pm$.010} & \best{2.710{\scriptsize$\pm$.019}} & 10.882{\scriptsize$\pm$.081} & 1.152{\scriptsize$\pm$.041} \\
\midrule
\textbf{\ours{} (ours)} & \second{0.440{\scriptsize$\pm$.007}} & \best{0.667{\scriptsize$\pm$.007}} & \second{0.790{\scriptsize$\pm$.006}} & \best{0.141{\scriptsize$\pm$.009}} & \second{2.720{\scriptsize$\pm$.026}} & 10.745{\scriptsize$\pm$.074} & \second{1.342{\scriptsize$\pm$.065}} \\
\bottomrule
\end{tabular}
\vspace{-15pt}
\end{table*}

Tables~\ref{tab:sota_h3d} and~\ref{tab:sota_kit} show a consistent picture.
Within the masked motion generation family, \ours{} achieves the best FID on both benchmarks.
On HumanML3D, it reduces FID from 0.045 (MoMask) to 0.028, a 38\% relative improvement, while simultaneously achieving the best MM-Dist (2.879) and Diversity (9.73).
On KIT-ML, it achieves the lowest FID (0.141) across all masked methods and competitive retrieval metrics.

The largest gains appear in FID and MM-Dist, the two metrics most directly related to motion realism and text-motion alignment.
This is consistent with the role of MSD: it reallocates modeling capacity toward dynamically difficult regions, improving motion fidelity rather than surface-level retrieval statistics.
In the broader cross-paradigm comparison, \ours{} remains competitive with strong recent methods~\cite{li2024lamp} that rely on auxiliary teachers or additional generative branches, while using only a simple, parameter-free complexity signal.
Relative to teacher-based hard-token methods~\cite{zhang2025echomask, jeonghgm3, wang2023hard}, our method has two practical advantages: it does not require an auxiliary teacher model, and it reuses the same motion-grounded signal beyond mask selection alone, including motion-aware attention in the core transformer.

\subsection{Ablation Study}
\label{sec:exp_ablation}

We ablate the two main design axes of \ours{}: (1)~which complexity signal to use, and (2)~how each proposed component contributes to the final result.

\noindent\textbf{Complexity signal choice.}
A natural question is what kind of signal should best represent local motion complexity.
Table~\ref{tab:signal} compares several alternatives under the same downstream setting.
Non-uniform treatment consistently outperforms purely uniform masking, confirming the general motivation for complexity-aware generation.
Among the tested signals, motion-grounded cues are more effective than the loss-based hardness proxy used in HGM3, suggesting that generation difficulty is better explained by motion-side structure than by optimization behavior.
MSD achieves the best FID because it provides both a scalar complexity summary for ranking difficult regions and a structured spectral profile for pairwise frame comparison in motion-aware attention---a dual utility that scalar cues cannot offer.

\begin{table}[t]
\centering
\caption{\textbf{Complexity signal comparison} on HumanML3D (mean $\pm$ 95\% CI).
MSD gives the strongest FID among the tested signals.}
\label{tab:signal}
\small
\setlength{\tabcolsep}{3pt}
\begin{tabular}{lccc}
\toprule
Complexity Signal & FID$\downarrow$ & Top-3$\uparrow$ & MM-Dist$\downarrow$ \\
\midrule
None (uniform) & {.045{\scriptsize$\pm$.003}} & {.807{\scriptsize$\pm$.004}} & {2.958{\scriptsize$\pm$.021}} \\
Velocity mag. & {.043{\scriptsize$\pm$.003}} & {.798{\scriptsize$\pm$.004}} & {3.016{\scriptsize$\pm$.020}} \\
Loss-based (HGM3) & {.042{\scriptsize$\pm$.002}} & {.792{\scriptsize$\pm$.004}} & {3.049{\scriptsize$\pm$.019}} \\
\msd{} + Loss & {.031{\scriptsize$\pm$.002}} & {.800{\scriptsize$\pm$.003}} & {3.013{\scriptsize$\pm$.017}} \\
\textbf{\msd{} (ours)} & \best{{.028{\scriptsize$\pm$.005}}} & {.795{\scriptsize$\pm$.003}} & \best{2.879{\scriptsize$\pm$.006}} \\
\bottomrule
\end{tabular}
\vspace{-15pt}
\end{table}

\begin{figure}[t]
\centering
\includegraphics[width=0.45\textwidth]{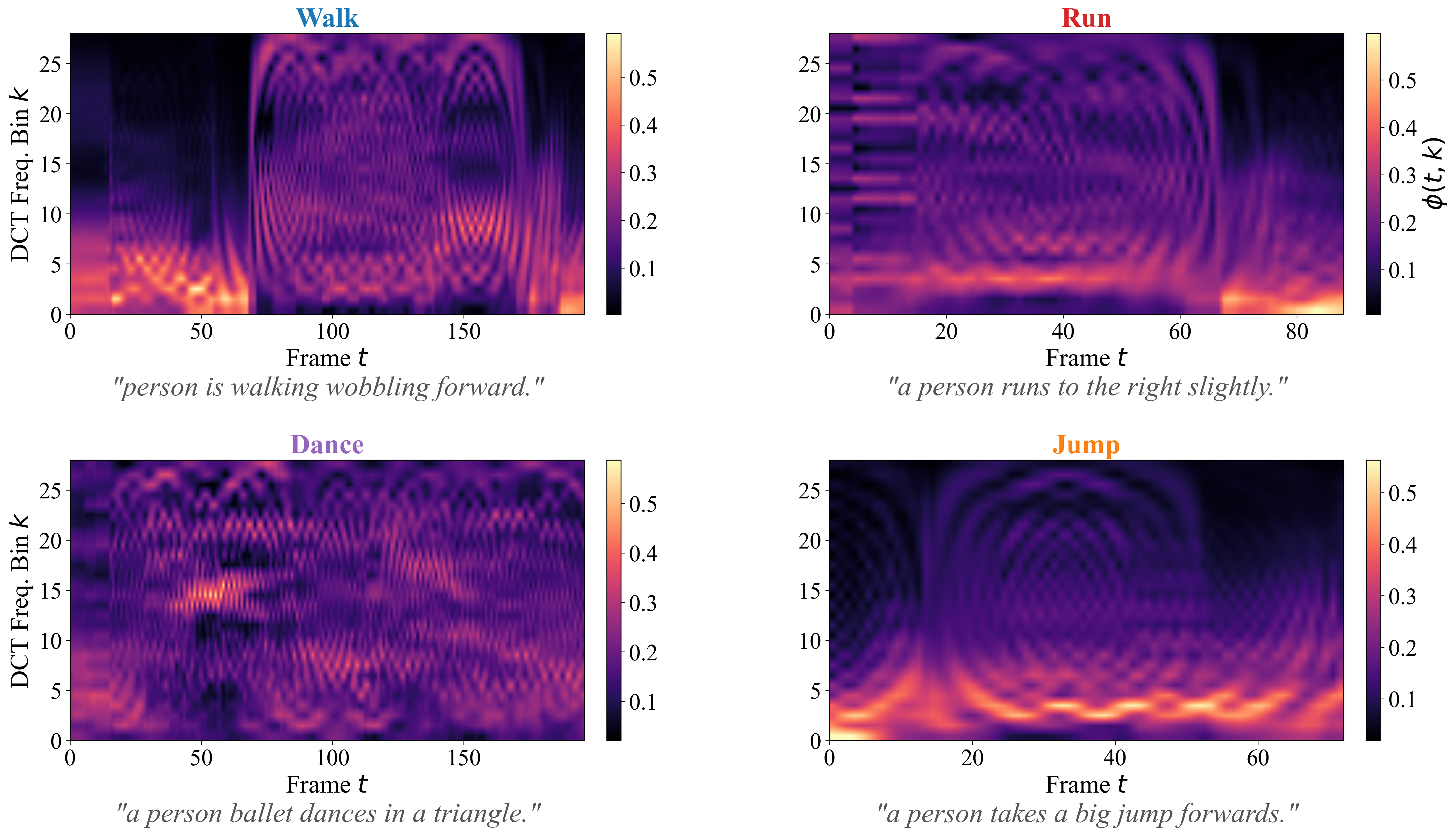}
\caption{\textbf{MSD spectral fingerprints.}
Representative MSD heatmaps for different motion types show distinct temporal-frequency patterns across time.}
\label{fig:msd_vis}
\vspace{-15pt}
\end{figure}

Fig.~\ref{fig:msd_vis} provides further intuition: representative MSD heatmaps exhibit clearly different spectral signatures across motion types, confirming that MSD captures meaningful local motion structure rather than an arbitrary training heuristic.

\noindent\textbf{Component contribution.}
We next study how each proposed component improves the masked pipeline, with a clear distinction between the \textbf{core model} (complexity-aware mask selection + motion-aware attention) and the \textbf{full model} (additionally including complexity-aware decoding at inference time).
Table~\ref{tab:additive} shows the additive ablation.
Starting from the masked baseline, spectral attention and complexity-aware masking each provide clear, non-redundant gains.
Mask selection contributes the larger improvement, which is expected because it directly reallocates reconstruction supervision toward dynamically difficult regions.
Adding complexity-aware decoding at inference time produces a further gain, confirming that local motion complexity is useful throughout the pipeline---not only for training-time mask allocation but also for decoding-time confidence estimation.

\begin{table}[t]
\centering
\caption{\textbf{Additive ablation} on HumanML3D (mean $\pm$ 95\% CI over 20 runs).
Each component provides a non-redundant gain.}
\label{tab:additive}
\small
\begin{tabular}{lccc}
\toprule
Configuration & FID$\downarrow$ & Top-3$\uparrow$ & MM-Dist$\downarrow$ \\
\midrule
DynMask (baseline) & {.045{\scriptsize$\pm$.003}} & {.807{\scriptsize$\pm$.004}} & {2.958{\scriptsize$\pm$.021}} \\
+ Spectral Attn (I) & {.039{\scriptsize$\pm$.002}} & {.802{\scriptsize$\pm$.004}} & {2.975{\scriptsize$\pm$.019}} \\
+ CFS Masking (II) & {.032{\scriptsize$\pm$.002}} & {.798{\scriptsize$\pm$.003}} & {3.010{\scriptsize$\pm$.018}} \\
+ DAS Sampling (III) & \best{{.028{\scriptsize$\pm$.005}}} & .795{\scriptsize$\pm$.003} & \best{2.879{\scriptsize$\pm$.006}} \\
\bottomrule
\end{tabular}
\vspace{-5pt}
\end{table}

Additional ablations in the appendix examine further implementation choices, including the input signal used to compute MSD, the fusion rule for motion-aware attention, the layer schedule for the spectral prior, simpler alternatives such as direct loss reweighting and feature concatenation, and the design of dual-quota mask allocation.

\subsection{Complexity-Stratified Analysis}
\label{sec:exp_complexity}

A central claim of this work is that complexity-aware generation should help most where motion is dynamically difficult.
We verify this claim through three complementary analyses.

\noindent\textbf{Per-category breakdown.}
We first divide HumanML3D test motions into static and dynamic groups by action semantics.
Table~\ref{tab:per_category} shows that the masked baseline degrades much more on the dynamic group (FID 0.227) than on the static group (FID 0.077), a 2.96$\times$ gap.
\ours{} reduces this gap to 1.77$\times$, confirming that complexity-aware generation is most helpful.

\begin{table}[t]
\centering
\caption{\textbf{Per-category FID} on HumanML3D.
Dynamic actions are substantially harder for the masked baseline, while \ours{} reduces the gap.
Motions are classified by keyword matching into static (3030 samples) and dynamic (1168 samples) groups.}
\label{tab:per_category}
\small
\begin{tabular}{lcc}
\toprule
Category & MoMask FID$\downarrow$ & \ours{} FID$\downarrow$ \\
\midrule
Static (walk, stand, wave, sit) & {0.077{\scriptsize$\pm$.002}} & {0.089{\scriptsize$\pm$.003}} \\
Dynamic (kick, spin, jump, dance) & {0.227{\scriptsize$\pm$.006}} & {0.158{\scriptsize$\pm$.004}} \\
\midrule
Ratio (dynamic / static) & {2.96$\times$} & {1.77$\times$} \\
\bottomrule
\end{tabular}
\vspace{-15pt}
\end{table}

\noindent\textbf{Per-complexity stratification.}
We further divide the test set into three groups by external physical criteria (motion dynamics and sequence length), independent of MSD itself.
Table~\ref{tab:complexity} shows that the gains concentrate on medium and complex groups ($-$16\% and $-$24\% FID), while the simple group sees a slight increase.
This trade-off is by design: \ours{} reallocates modeling effort toward dynamically difficult regions, which may come at a small cost on the simplest motions where uniform masking is already sufficient.
Crucially, the overall FID still improves because the complex-motion gains substantially outweigh the simple-motion cost, and the net effect is positive across the full test set (Table~\ref{tab:sota_h3d}).
The same trend appears in the length-stratified analysis in the appendix.

\begin{table}[t]
\centering
\caption{\textbf{Per-complexity FID} on HumanML3D.
Subgroup FID is higher than full-set FID because each group contains fewer samples.}
\label{tab:complexity}
\small
\begin{tabular}{lccc}
\toprule
Complexity & MoMask & \ours{} & $\Delta$FID \\
\midrule
Simple & {0.440{\scriptsize$\pm$.007}} & {0.560{\scriptsize$\pm$.010}} & {$+$27\%} \\
Medium & {0.221{\scriptsize$\pm$.006}} & {0.186{\scriptsize$\pm$.003}} & {$-$16\%} \\
Complex & {0.604{\scriptsize$\pm$.010}} & {0.462{\scriptsize$\pm$.010}} & \textbf{{$-$24\%}} \\
\bottomrule
\end{tabular}
\vspace{-15pt}
\end{table}

\noindent\textbf{Cross-dataset transfer.}
To test whether the benefit of MSD generalizes beyond the training distribution, we evaluate a HumanML3D-trained model zero-shot on BABEL, a dataset with different action vocabulary and annotation style.
As shown in Table~\ref{tab:babel}, \ours{} reduces FID from {14.889} to {11.543}, a relative improvement of {22.5\%}.
This is notable because MSD is computed entirely from motion kinematics without any dataset-specific tuning, suggesting that the complexity prior captures general physical motion structure rather than distributional artifacts of a particular dataset.

\begin{table}[t]
\centering
\caption{\textbf{Cross-dataset transfer} (HumanML3D $\to$ BABEL, zero-shot).}
\label{tab:babel}
\small
\begin{tabular}{lcc}
\toprule
Model & FID$\downarrow$ & $\Delta$ \\
\midrule
MoMask & {14.889{\scriptsize$\pm$.038}} & --- \\
\textbf{\ours{}} & \best{{11.543{\scriptsize$\pm$.040}}} & {$-$22.5\%} \\
\bottomrule
\end{tabular}
\vspace{-15pt}
\end{table}

\subsection{Qualitative Evaluation and User Study}
\label{sec:exp_user}

\begin{figure*}[t]
\centering
\includegraphics[width=0.76\textwidth]{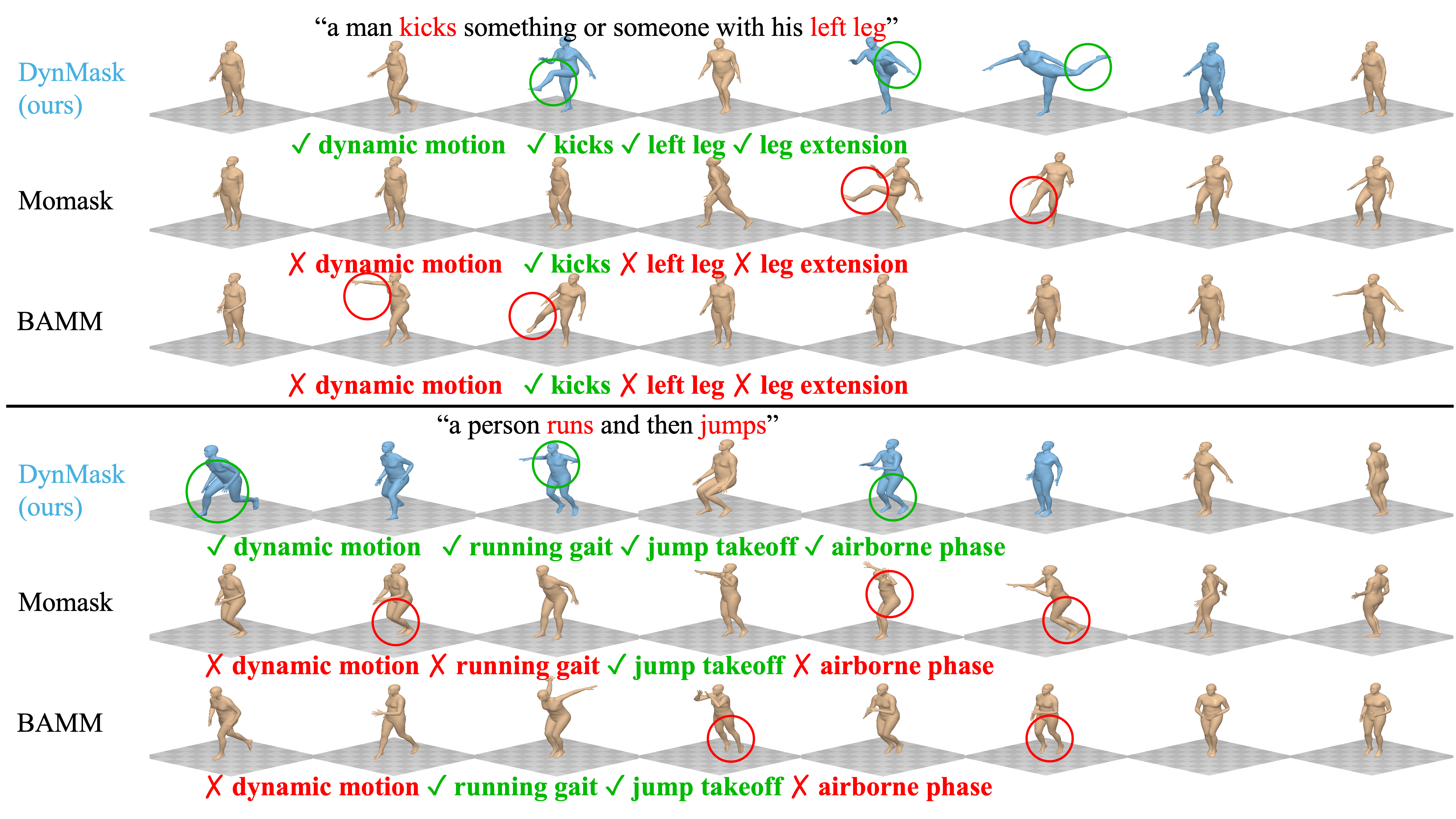}
\caption{\textbf{Qualitative comparison on challenging dynamic prompts.}
Compared with representative masked baselines, \ours{} better preserves dynamic timing, side-specific limb actions, kick extension, running gait, jump takeoff, and airborne phases.
Green circles highlight locally correct dynamic details, while red circles mark failure cases in the baselines.}
\label{fig:qualitative}
\vspace{-15pt}
\end{figure*}
\begin{figure*}[t]
\centering
\includegraphics[width=0.85\textwidth]{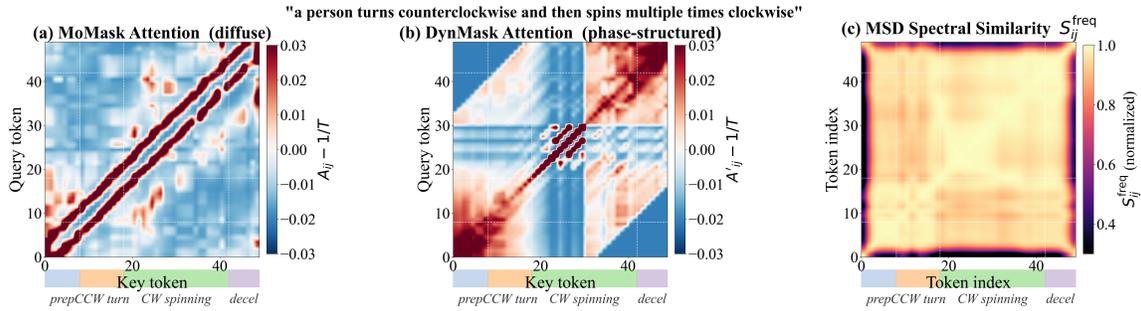}
\caption{\textbf{Attention visualization on a multi-phase turning-and-spinning sequence.}
\textbf{(a)}~The MoMask baseline produces comparatively diffuse attention and weak phase separation.
\textbf{(b)}~\ours{} yields clearer phase-structured attention aligned with the underlying motion segments.
\textbf{(c)}~The MSD spectral similarity matrix shows strong within-phase consistency, explaining why motion-aware attention connects dynamically compatible regions more effectively.}
\label{fig:attn_vis}
\vspace{-15pt}
\end{figure*}
\noindent\textbf{Qualitative comparison.}
Fig.~\ref{fig:qualitative} shows representative prompts with strong local dynamics and fine-grained semantic constraints.
Compared with strong masked baselines, \ours{} better preserves local phase continuity, dynamic timing, and motion-specific details such as kick extension, side-specific limb control, running gait, jump takeoff, and airborne phases.
In the first example, \ours{} not only captures the kicking action itself, but also preserves the correct left-leg execution and full leg extension, whereas the baselines either weaken the dynamic motion or miss the limb-specific semantic constraint.
In the second example, \ours{} generates a clearer transition from running to jumping, with more plausible takeoff preparation and a more distinct airborne stage, while the baselines tend to produce weaker dynamics or collapse the motion into a less structured sequence.
These qualitative differences are consistent with our main claim: complexity-aware generation is most useful in motion regions where dynamics change rapidly and where standard masked generation tends to under-model critical local structure.

\noindent \textbf{User study.}
We recruit 37 participants to rank blind-shuffled videos from \ours{}, MoMask, and BAMM along three criteria: motion quality, text faithfulness, and temporal naturalness.
Table~\ref{tab:user_study} reports the percentage ranked first.
\ours{} is consistently preferred across all criteria, with the largest margin on text faithfulness (67.9\% overall, 64.6\% on complex motions).
The consistent ordering \ours{} $>$ MoMask $>$ BAMM aligns with the quantitative results.

\begin{table}[t]
\centering
\caption{\textbf{User study} (\% ranked 1st).
\ours{} is preferred across across all criteria.}
\label{tab:user_study}
\small
\setlength{\tabcolsep}{3pt}
\resizebox{0.45\textwidth}{!}{\begin{tabular}{lccc|ccc}
\toprule
\multirow{2}{*}{Criterion} & \multicolumn{3}{c|}{Overall} & \multicolumn{3}{c}{Complex only} \\
 & \ours{} & MoMask & BAMM & \ours{} & MoMask & BAMM \\
\midrule
Motion Quality & \best{55.4} & 26.8 & 21.4 & \best{45.8} & 31.2 & 25.0\\
Text Faithfulness & \best{67.9} & 23.2 & 15.2 & \best{64.6} & 27.1 & 10.4 \\
Temporal Naturalness & \best{53.6} & 25.9 & 25.0 & \best{45.8} & 29.2 & 25.0\\
\bottomrule
\end{tabular}}
\vspace{-15pt}
\end{table}

\subsection{Further Analysis}
\label{sec:exp_more}

\noindent\textbf{Attention structure.}
Fig.~\ref{fig:attn_vis} visualizes attention patterns for a multi-phase spinning sequence.
MoMask's attention is diffuse with no clear motion-phase structure, meaning that frames from different dynamic phases attend to each other indiscriminately.
In contrast, \ours{}'s spectral attention bias induces clear block-diagonal structure aligned with motion phases: frames within the same dynamic phase attend strongly to each other, while cross-phase attention is suppressed.
The MSD spectral similarity matrix $S^{\text{freq}}$ (Fig.~\ref{fig:attn_vis}c) explains this structure---frames with similar spectral profiles naturally form coherent blocks.
This supports our claim that MSD is useful not only for complexity-aware mask allocation, but also for organizing long-range information flow in the transformer, helping the model reason about motion dynamics in a phase-aware manner.

\noindent\textbf{Sensitivity, computation, and editing.}
We vary the main hyperparameters over broad ranges and find that performance remains stable around the default setting, indicating that \ours{} does not require careful tuning.
MSD introduces no additional trainable parameters, while the current implementation adds moderate training overhead and higher inference-time cost due to iterative spectral recomputation; detailed timing is reported in the appendix.
Because the method builds on masked generation, it also naturally supports temporal inpainting and prefix completion, where we observe consistent gains in the appendix.

\section{Conclusion}

\label{sec:conclusion}

\label{sec:conclusion}

We have presented \ours{}, a complexity-aware masked motion generation framework that addresses a fundamental mismatch in existing pipelines: motion complexity varies sharply over time, yet standard masked models treat all frames uniformly during both training and inference.
At the core of our approach is the Motion Spectral Descriptor (\msd{}), a parameter-free signal derived from the short-time spectrum of motion velocity that provides a practical estimate of local dynamic complexity.
By integrating MSD into mask selection, attention, and decoding, \ours{} achieves state-of-the-art FID within the masked motion generation family on both HumanML3D and KIT-ML, with the largest gains concentrated on dynamically complex motions.
Complexity-stratified evaluation and cross-dataset transfer experiments confirm that the improvements are consistent and generalizable, and a user study validates that these gains translate into clear perceptual preferences.

\bibliographystyle{ACM-Reference-Format}

\begin{thebibliography}{76}


\ifx \showCODEN    \undefined \def \showCODEN     #1{\unskip}     \fi
\ifx \showISBNx    \undefined \def \showISBNx     #1{\unskip}     \fi
\ifx \showISBNxiii \undefined \def \showISBNxiii  #1{\unskip}     \fi
\ifx \showISSN     \undefined \def \showISSN      #1{\unskip}     \fi
\ifx \showLCCN     \undefined \def \showLCCN      #1{\unskip}     \fi
\ifx \shownote     \undefined \def \shownote      #1{#1}          \fi
\ifx \showarticletitle \undefined \def \showarticletitle #1{#1}   \fi
\ifx \showURL      \undefined \def \showURL       {\relax}        \fi
\providecommand\bibfield[2]{#2}
\providecommand\bibinfo[2]{#2}
\providecommand\natexlab[1]{#1}
\providecommand\showeprint[2][]{arXiv:#2}

\bibitem[Ahmed et~al\mbox{.}(1974)]%
        {ahmed1974discrete}
\bibfield{author}{\bibinfo{person}{Nasir Ahmed}, \bibinfo{person}{T\_ Natarajan}, {and} \bibinfo{person}{Kamisetty~R Rao}.} \bibinfo{year}{1974}\natexlab{}.
\newblock \showarticletitle{Discrete cosine transform}.
\newblock \bibinfo{journal}{\emph{IEEE transactions on Computers}} \bibinfo{volume}{100}, \bibinfo{number}{1} (\bibinfo{year}{1974}), \bibinfo{pages}{90--93}.
\newblock


\bibitem[Ao et~al\mbox{.}(2022)]%
        {ao2022rhythmic}
\bibfield{author}{\bibinfo{person}{Tenglong Ao}, \bibinfo{person}{Qingzhe Gao}, \bibinfo{person}{Yuke Lou}, \bibinfo{person}{Baoquan Chen}, {and} \bibinfo{person}{Libin Liu}.} \bibinfo{year}{2022}\natexlab{}.
\newblock \showarticletitle{Rhythmic gesticulator: Rhythm-aware co-speech gesture synthesis with hierarchical neural embeddings}.
\newblock \bibinfo{journal}{\emph{ACM Transactions on Graphics (TOG)}} \bibinfo{volume}{41}, \bibinfo{number}{6} (\bibinfo{year}{2022}), \bibinfo{pages}{1--19}.
\newblock


\bibitem[Ao et~al\mbox{.}(2023)]%
        {ao2023gesturediffuclip}
\bibfield{author}{\bibinfo{person}{Tenglong Ao}, \bibinfo{person}{Zeyi Zhang}, {and} \bibinfo{person}{Libin Liu}.} \bibinfo{year}{2023}\natexlab{}.
\newblock \showarticletitle{Gesturediffuclip: Gesture diffusion model with clip latents}.
\newblock \bibinfo{journal}{\emph{ACM Transactions on Graphics (TOG)}} \bibinfo{volume}{42}, \bibinfo{number}{4} (\bibinfo{year}{2023}), \bibinfo{pages}{1--18}.
\newblock


\bibitem[Bandara et~al\mbox{.}(2023)]%
        {bandara2023adamae}
\bibfield{author}{\bibinfo{person}{Wele Gedara~Chaminda Bandara}, \bibinfo{person}{Naman Patel}, \bibinfo{person}{Ali Gholami}, \bibinfo{person}{Mehdi Nikkhah}, \bibinfo{person}{Motilal Agrawal}, {and} \bibinfo{person}{Vishal~M Patel}.} \bibinfo{year}{2023}\natexlab{}.
\newblock \showarticletitle{Adamae: Adaptive masking for efficient spatiotemporal learning with masked autoencoders}. In \bibinfo{booktitle}{\emph{Proceedings of the IEEE/CVF Conference on Computer Vision and Pattern Recognition}}. \bibinfo{pages}{14507--14517}.
\newblock


\bibitem[Bao et~al\mbox{.}(2021)]%
        {bao2021beit}
\bibfield{author}{\bibinfo{person}{Hangbo Bao}, \bibinfo{person}{Li Dong}, \bibinfo{person}{Songhao Piao}, {and} \bibinfo{person}{Furu Wei}.} \bibinfo{year}{2021}\natexlab{}.
\newblock \showarticletitle{Beit: Bert pre-training of image transformers}.
\newblock \bibinfo{journal}{\emph{arXiv preprint arXiv:2106.08254}} (\bibinfo{year}{2021}).
\newblock


\bibitem[Chang et~al\mbox{.}(2023)]%
        {chang2023muse}
\bibfield{author}{\bibinfo{person}{Huiwen Chang}, \bibinfo{person}{Han Zhang}, \bibinfo{person}{Jarred Barber}, \bibinfo{person}{AJ Maschinot}, \bibinfo{person}{Jose Lezama}, \bibinfo{person}{Lu Jiang}, \bibinfo{person}{Ming-Hsuan Yang}, \bibinfo{person}{Kevin Murphy}, \bibinfo{person}{William~T Freeman}, \bibinfo{person}{Michael Rubinstein}, {et~al\mbox{.}}} \bibinfo{year}{2023}\natexlab{}.
\newblock \showarticletitle{Muse: Text-to-image generation via masked generative transformers}.
\newblock \bibinfo{journal}{\emph{arXiv preprint arXiv:2301.00704}} (\bibinfo{year}{2023}).
\newblock


\bibitem[Chang et~al\mbox{.}(2022)]%
        {chang2022maskgit}
\bibfield{author}{\bibinfo{person}{Huiwen Chang}, \bibinfo{person}{Han Zhang}, \bibinfo{person}{Lu Jiang}, \bibinfo{person}{Ce Liu}, {and} \bibinfo{person}{William~T Freeman}.} \bibinfo{year}{2022}\natexlab{}.
\newblock \showarticletitle{Maskgit: Masked generative image transformer}. In \bibinfo{booktitle}{\emph{Proceedings of the IEEE/CVF conference on computer vision and pattern recognition}}. \bibinfo{pages}{11315--11325}.
\newblock


\bibitem[Chen et~al\mbox{.}(2023)]%
        {chen2023executing}
\bibfield{author}{\bibinfo{person}{Xin Chen}, \bibinfo{person}{Biao Jiang}, \bibinfo{person}{Wen Liu}, \bibinfo{person}{Zilong Huang}, \bibinfo{person}{Bin Fu}, \bibinfo{person}{Tao Chen}, {and} \bibinfo{person}{Gang Yu}.} \bibinfo{year}{2023}\natexlab{}.
\newblock \showarticletitle{Executing your commands via motion diffusion in latent space}. In \bibinfo{booktitle}{\emph{Proceedings of the IEEE/CVF conference on computer vision and pattern recognition}}. \bibinfo{pages}{18000--18010}.
\newblock


\bibitem[Devlin et~al\mbox{.}(2019)]%
        {devlin2019bert}
\bibfield{author}{\bibinfo{person}{Jacob Devlin}, \bibinfo{person}{Ming-Wei Chang}, \bibinfo{person}{Kenton Lee}, {and} \bibinfo{person}{Kristina Toutanova}.} \bibinfo{year}{2019}\natexlab{}.
\newblock \showarticletitle{Bert: Pre-training of deep bidirectional transformers for language understanding}. In \bibinfo{booktitle}{\emph{Proceedings of the 2019 conference of the North American chapter of the association for computational linguistics: human language technologies, volume 1 (long and short papers)}}. \bibinfo{pages}{4171--4186}.
\newblock


\bibitem[Dosovitskiy et~al\mbox{.}(2020)]%
        {dosovitskiy2020image}
\bibfield{author}{\bibinfo{person}{Alexey Dosovitskiy}, \bibinfo{person}{Lucas Beyer}, \bibinfo{person}{Alexander Kolesnikov}, \bibinfo{person}{Dirk Weissenborn}, \bibinfo{person}{Xiaohua Zhai}, \bibinfo{person}{Thomas Unterthiner}, \bibinfo{person}{Mostafa Dehghani}, \bibinfo{person}{Matthias Minderer}, \bibinfo{person}{Georg Heigold}, \bibinfo{person}{Sylvain Gelly}, {et~al\mbox{.}}} \bibinfo{year}{2020}\natexlab{}.
\newblock \showarticletitle{An image is worth 16x16 words: Transformers for image recognition at scale}.
\newblock \bibinfo{journal}{\emph{arXiv preprint arXiv:2010.11929}} (\bibinfo{year}{2020}).
\newblock


\bibitem[Guo et~al\mbox{.}(2024)]%
        {guo2024momask}
\bibfield{author}{\bibinfo{person}{Chuan Guo}, \bibinfo{person}{Yuxuan Mu}, \bibinfo{person}{Muhammad~Gohar Javed}, \bibinfo{person}{Sen Wang}, {and} \bibinfo{person}{Li Cheng}.} \bibinfo{year}{2024}\natexlab{}.
\newblock \showarticletitle{Momask: Generative masked modeling of 3d human motions}. In \bibinfo{booktitle}{\emph{Proceedings of the IEEE/CVF Conference on Computer Vision and Pattern Recognition}}. \bibinfo{pages}{1900--1910}.
\newblock


\bibitem[Guo et~al\mbox{.}(2022)]%
        {guo2022generating}
\bibfield{author}{\bibinfo{person}{Chuan Guo}, \bibinfo{person}{Shihao Zou}, \bibinfo{person}{Xinxin Zuo}, \bibinfo{person}{Sen Wang}, \bibinfo{person}{Wei Ji}, \bibinfo{person}{Xingyu Li}, {and} \bibinfo{person}{Li Cheng}.} \bibinfo{year}{2022}\natexlab{}.
\newblock \showarticletitle{Generating diverse and natural 3d human motions from text}. In \bibinfo{booktitle}{\emph{Proceedings of the IEEE/CVF conference on computer vision and pattern recognition}}. \bibinfo{pages}{5152--5161}.
\newblock


\bibitem[Hou et~al\mbox{.}(2022)]%
        {hou2022milan}
\bibfield{author}{\bibinfo{person}{Zejiang Hou}, \bibinfo{person}{Fei Sun}, \bibinfo{person}{Yen-Kuang Chen}, \bibinfo{person}{Yuan Xie}, {and} \bibinfo{person}{Sun-Yuan Kung}.} \bibinfo{year}{2022}\natexlab{}.
\newblock \showarticletitle{Milan: Masked image pretraining on language assisted representation}.
\newblock \bibinfo{journal}{\emph{arXiv preprint arXiv:2208.06049}} (\bibinfo{year}{2022}).
\newblock


\bibitem[Jeong et~al\mbox{.}({[n.\,d.]})]%
        {jeonghgm3}
\bibfield{author}{\bibinfo{person}{Minjae Jeong}, \bibinfo{person}{Yechan Hwang}, \bibinfo{person}{Jaejin Lee}, \bibinfo{person}{Sungyoon Jung}, {and} \bibinfo{person}{Won~Hwa Kim}.} \bibinfo{year}{[n.\,d.]}\natexlab{}.
\newblock \showarticletitle{HGM$^3$: Hierarchical Generative Masked Motion Modeling with Hard Token Mining}. In \bibinfo{booktitle}{\emph{The Thirteenth International Conference on Learning Representations}}.
\newblock


\bibitem[Jiang et~al\mbox{.}(2023)]%
        {jiang2023motiongpt}
\bibfield{author}{\bibinfo{person}{Biao Jiang}, \bibinfo{person}{Xin Chen}, \bibinfo{person}{Wen Liu}, \bibinfo{person}{Jingyi Yu}, \bibinfo{person}{Gang Yu}, {and} \bibinfo{person}{Tao Chen}.} \bibinfo{year}{2023}\natexlab{}.
\newblock \showarticletitle{Motiongpt: Human motion as a foreign language}.
\newblock \bibinfo{journal}{\emph{Advances in Neural Information Processing Systems}}  \bibinfo{volume}{36} (\bibinfo{year}{2023}), \bibinfo{pages}{20067--20079}.
\newblock


\bibitem[Kakogeorgiou et~al\mbox{.}(2022)]%
        {kakogeorgiou2022hide}
\bibfield{author}{\bibinfo{person}{Ioannis Kakogeorgiou}, \bibinfo{person}{Spyros Gidaris}, \bibinfo{person}{Bill Psomas}, \bibinfo{person}{Yannis Avrithis}, \bibinfo{person}{Andrei Bursuc}, \bibinfo{person}{Konstantinos Karantzalos}, {and} \bibinfo{person}{Nikos Komodakis}.} \bibinfo{year}{2022}\natexlab{}.
\newblock \showarticletitle{What to hide from your students: Attention-guided masked image modeling}. In \bibinfo{booktitle}{\emph{European Conference on Computer Vision}}. Springer, \bibinfo{pages}{300--318}.
\newblock


\bibitem[Karunratanakul et~al\mbox{.}(2023)]%
        {karunratanakul2023guided}
\bibfield{author}{\bibinfo{person}{Korrawe Karunratanakul}, \bibinfo{person}{Konpat Preechakul}, \bibinfo{person}{Supasorn Suwajanakorn}, {and} \bibinfo{person}{Siyu Tang}.} \bibinfo{year}{2023}\natexlab{}.
\newblock \showarticletitle{Guided motion diffusion for controllable human motion synthesis}. In \bibinfo{booktitle}{\emph{Proceedings of the IEEE/CVF international conference on computer vision}}. \bibinfo{pages}{2151--2162}.
\newblock


\bibitem[Li et~al\mbox{.}(2026)]%
        {li2025frankenmotion}
\bibfield{author}{\bibinfo{person}{Chuqiao Li}, \bibinfo{person}{Xianghui Xie}, \bibinfo{person}{Yong Cao}, \bibinfo{person}{Andreas Geiger}, {and} \bibinfo{person}{Gerard Pons-Moll}.} \bibinfo{year}{2026}\natexlab{}.
\newblock \showarticletitle{FrankenMotion: Part-level Human Motion Generation and Composition}.
\newblock \bibinfo{journal}{\emph{arXiv preprint arXiv:2601.10909}} (\bibinfo{year}{2026}).
\newblock


\bibitem[Li et~al\mbox{.}(2024b)]%
        {li2024lodge}
\bibfield{author}{\bibinfo{person}{Ronghui Li}, \bibinfo{person}{YuXiang Zhang}, \bibinfo{person}{Yachao Zhang}, \bibinfo{person}{Hongwen Zhang}, \bibinfo{person}{Jie Guo}, \bibinfo{person}{Yan Zhang}, \bibinfo{person}{Yebin Liu}, {and} \bibinfo{person}{Xiu Li}.} \bibinfo{year}{2024}\natexlab{b}.
\newblock \showarticletitle{Lodge: A coarse to fine diffusion network for long dance generation guided by the characteristic dance primitives}. In \bibinfo{booktitle}{\emph{Proceedings of the IEEE/CVF Conference on Computer Vision and Pattern Recognition}}. \bibinfo{pages}{1524--1534}.
\newblock


\bibitem[Li et~al\mbox{.}(2023)]%
        {li2023finedance}
\bibfield{author}{\bibinfo{person}{Ronghui Li}, \bibinfo{person}{Junfan Zhao}, \bibinfo{person}{Yachao Zhang}, \bibinfo{person}{Mingyang Su}, \bibinfo{person}{Zeping Ren}, \bibinfo{person}{Han Zhang}, \bibinfo{person}{Yansong Tang}, {and} \bibinfo{person}{Xiu Li}.} \bibinfo{year}{2023}\natexlab{}.
\newblock \showarticletitle{Finedance: A fine-grained choreography dataset for 3d full body dance generation}. In \bibinfo{booktitle}{\emph{Proceedings of the IEEE/CVF International Conference on Computer Vision}}. \bibinfo{pages}{10234--10243}.
\newblock


\bibitem[Li et~al\mbox{.}(2024a)]%
        {li2024lamp}
\bibfield{author}{\bibinfo{person}{Zhe Li}, \bibinfo{person}{Weihao Yuan}, \bibinfo{person}{Yisheng He}, \bibinfo{person}{Lingteng Qiu}, \bibinfo{person}{Shenhao Zhu}, \bibinfo{person}{Xiaodong Gu}, \bibinfo{person}{Weichao Shen}, \bibinfo{person}{Yuan Dong}, \bibinfo{person}{Zilong Dong}, {and} \bibinfo{person}{Laurence~T Yang}.} \bibinfo{year}{2024}\natexlab{a}.
\newblock \showarticletitle{LaMP: Language-Motion Pretraining for Motion Generation, Retrieval, and Captioning}.
\newblock \bibinfo{journal}{\emph{arXiv preprint arXiv:2410.07093}} (\bibinfo{year}{2024}).
\newblock


\bibitem[Liu et~al\mbox{.}(2024)]%
        {liu2024emage}
\bibfield{author}{\bibinfo{person}{Haiyang Liu}, \bibinfo{person}{Zihao Zhu}, \bibinfo{person}{Giorgio Becherini}, \bibinfo{person}{Yichen Peng}, \bibinfo{person}{Mingyang Su}, \bibinfo{person}{You Zhou}, \bibinfo{person}{Xuefei Zhe}, \bibinfo{person}{Naoya Iwamoto}, \bibinfo{person}{Bo Zheng}, {and} \bibinfo{person}{Michael~J Black}.} \bibinfo{year}{2024}\natexlab{}.
\newblock \showarticletitle{EMAGE: Towards unified holistic co-speech gesture generation via expressive masked audio gesture modeling}. In \bibinfo{booktitle}{\emph{Proceedings of the IEEE/CVF Conference on Computer Vision and Pattern Recognition}}. \bibinfo{pages}{1144--1154}.
\newblock


\bibitem[Liu et~al\mbox{.}(2022)]%
        {liu2022beat}
\bibfield{author}{\bibinfo{person}{Haiyang Liu}, \bibinfo{person}{Zihao Zhu}, \bibinfo{person}{Naoya Iwamoto}, \bibinfo{person}{Yichen Peng}, \bibinfo{person}{Zhengqing Li}, \bibinfo{person}{You Zhou}, \bibinfo{person}{Elif Bozkurt}, {and} \bibinfo{person}{Bo Zheng}.} \bibinfo{year}{2022}\natexlab{}.
\newblock \showarticletitle{Beat: A large-scale semantic and emotional multi-modal dataset for conversational gestures synthesis}. In \bibinfo{booktitle}{\emph{European conference on computer vision}}. Springer, \bibinfo{pages}{612--630}.
\newblock


\bibitem[Liu et~al\mbox{.}(2023)]%
        {liu2023good}
\bibfield{author}{\bibinfo{person}{Zhengqi Liu}, \bibinfo{person}{Jie Gui}, {and} \bibinfo{person}{Hao Luo}.} \bibinfo{year}{2023}\natexlab{}.
\newblock \showarticletitle{Good helper is around you: Attention-driven masked image modeling}. In \bibinfo{booktitle}{\emph{Proceedings of the AAAI Conference on Artificial Intelligence}}, Vol.~\bibinfo{volume}{37}. \bibinfo{pages}{1799--1807}.
\newblock


\bibitem[Lou et~al\mbox{.}(2023)]%
        {lou2023diversemotion}
\bibfield{author}{\bibinfo{person}{Yunhong Lou}, \bibinfo{person}{Linchao Zhu}, \bibinfo{person}{Yaxiong Wang}, \bibinfo{person}{Xiaohan Wang}, {and} \bibinfo{person}{Yi Yang}.} \bibinfo{year}{2023}\natexlab{}.
\newblock \showarticletitle{Diversemotion: Towards diverse human motion generation via discrete diffusion}.
\newblock \bibinfo{journal}{\emph{arXiv preprint arXiv:2309.01372}} (\bibinfo{year}{2023}).
\newblock


\bibitem[Ma et~al\mbox{.}(2025)]%
        {ma2025intersyn}
\bibfield{author}{\bibinfo{person}{Yiyi Ma}, \bibinfo{person}{Yuanzhi Liang}, \bibinfo{person}{Xiu Li}, \bibinfo{person}{Chi Zhang}, {and} \bibinfo{person}{Xuelong Li}.} \bibinfo{year}{2025}\natexlab{}.
\newblock \showarticletitle{Intersyn: Interleaved learning for dynamic motion synthesis in the wild}. In \bibinfo{booktitle}{\emph{Proceedings of the IEEE/CVF International Conference on Computer Vision}}. \bibinfo{pages}{12832--12841}.
\newblock


\bibitem[Pan et~al\mbox{.}(2025)]%
        {pan2025tokenhsi}
\bibfield{author}{\bibinfo{person}{Liang Pan}, \bibinfo{person}{Zeshi Yang}, \bibinfo{person}{Zhiyang Dou}, \bibinfo{person}{Wenjia Wang}, \bibinfo{person}{Buzhen Huang}, \bibinfo{person}{Bo Dai}, \bibinfo{person}{Taku Komura}, {and} \bibinfo{person}{Jingbo Wang}.} \bibinfo{year}{2025}\natexlab{}.
\newblock \showarticletitle{Tokenhsi: Unified synthesis of physical human-scene interactions through task tokenization}. In \bibinfo{booktitle}{\emph{Proceedings of the Computer Vision and Pattern Recognition Conference}}. \bibinfo{pages}{5379--5391}.
\newblock


\bibitem[Petrovich et~al\mbox{.}(2022)]%
        {petrovich2022temos}
\bibfield{author}{\bibinfo{person}{Mathis Petrovich}, \bibinfo{person}{Michael~J Black}, {and} \bibinfo{person}{G{\"u}l Varol}.} \bibinfo{year}{2022}\natexlab{}.
\newblock \showarticletitle{Temos: Generating diverse human motions from textual descriptions}. In \bibinfo{booktitle}{\emph{European conference on computer vision}}. Springer, \bibinfo{pages}{480--497}.
\newblock


\bibitem[Petrovich et~al\mbox{.}(2024)]%
        {petrovich2024stmc}
\bibfield{author}{\bibinfo{person}{Mathis Petrovich}, \bibinfo{person}{Or Litany}, \bibinfo{person}{Umar Iqbal}, \bibinfo{person}{Michael~J Black}, \bibinfo{person}{Gul Varol}, \bibinfo{person}{Xue Bin~Peng}, {and} \bibinfo{person}{Davis Rempe}.} \bibinfo{year}{2024}\natexlab{}.
\newblock \showarticletitle{Multi-track timeline control for text-driven 3d human motion generation}. In \bibinfo{booktitle}{\emph{Proceedings of the IEEE/CVF Conference on Computer Vision and Pattern Recognition}}. \bibinfo{pages}{1911--1921}.
\newblock


\bibitem[Pinyoanuntapong et~al\mbox{.}(2024a)]%
        {pinyoanuntapong2024bamm}
\bibfield{author}{\bibinfo{person}{Ekkasit Pinyoanuntapong}, \bibinfo{person}{Muhammad~Usama Saleem}, \bibinfo{person}{Pu Wang}, \bibinfo{person}{Minwoo Lee}, \bibinfo{person}{Srijan Das}, {and} \bibinfo{person}{Chen Chen}.} \bibinfo{year}{2024}\natexlab{a}.
\newblock \showarticletitle{BAMM: bidirectional autoregressive motion model}. In \bibinfo{booktitle}{\emph{European Conference on Computer Vision}}. Springer, \bibinfo{pages}{172--190}.
\newblock


\bibitem[Pinyoanuntapong et~al\mbox{.}(2024b)]%
        {pinyoanuntapong2024mmm}
\bibfield{author}{\bibinfo{person}{Ekkasit Pinyoanuntapong}, \bibinfo{person}{Pu Wang}, \bibinfo{person}{Minwoo Lee}, {and} \bibinfo{person}{Chen Chen}.} \bibinfo{year}{2024}\natexlab{b}.
\newblock \showarticletitle{Mmm: Generative masked motion model}. In \bibinfo{booktitle}{\emph{Proceedings of the IEEE/CVF Conference on Computer Vision and Pattern Recognition}}. \bibinfo{pages}{1546--1555}.
\newblock


\bibitem[Plappert et~al\mbox{.}(2016)]%
        {plappert2016kit}
\bibfield{author}{\bibinfo{person}{Matthias Plappert}, \bibinfo{person}{Christian Mandery}, {and} \bibinfo{person}{Tamim Asfour}.} \bibinfo{year}{2016}\natexlab{}.
\newblock \showarticletitle{The kit motion-language dataset}.
\newblock \bibinfo{journal}{\emph{Big data}} \bibinfo{volume}{4}, \bibinfo{number}{4} (\bibinfo{year}{2016}), \bibinfo{pages}{236--252}.
\newblock


\bibitem[Punnakkal et~al\mbox{.}(2021)]%
        {punnakkal2021babel}
\bibfield{author}{\bibinfo{person}{Abhinanda~R Punnakkal}, \bibinfo{person}{Arjun Chandrasekaran}, \bibinfo{person}{Nikos Athanasiou}, \bibinfo{person}{Alejandra Quiros-Ramirez}, {and} \bibinfo{person}{Michael~J Black}.} \bibinfo{year}{2021}\natexlab{}.
\newblock \showarticletitle{BABEL: Bodies, action and behavior with english labels}. In \bibinfo{booktitle}{\emph{Proceedings of the IEEE/CVF conference on computer vision and pattern recognition}}. \bibinfo{pages}{722--731}.
\newblock


\bibitem[Ruiz-Ponce et~al\mbox{.}(2024)]%
        {ruizponce2024in2in}
\bibfield{author}{\bibinfo{person}{Pablo Ruiz-Ponce}, \bibinfo{person}{German Barquero}, \bibinfo{person}{Cristina Palmero}, \bibinfo{person}{Sergio Escalera}, {and} \bibinfo{person}{Jos{\'e} Garc{\'\i}a-Rodr{\'\i}guez}.} \bibinfo{year}{2024}\natexlab{}.
\newblock \showarticletitle{in2in: Leveraging individual information to generate human interactions}. In \bibinfo{booktitle}{\emph{Proceedings of the IEEE/CVF Conference on Computer Vision and Pattern Recognition}}. \bibinfo{pages}{1941--1951}.
\newblock


\bibitem[Ruiz-Ponce et~al\mbox{.}(2025)]%
        {ruizponce2025mixermdm}
\bibfield{author}{\bibinfo{person}{Pablo Ruiz-Ponce}, \bibinfo{person}{German Barquero}, \bibinfo{person}{Cristina Palmero}, \bibinfo{person}{Sergio Escalera}, {and} \bibinfo{person}{Jos{\'e} Garc{\'\i}a-Rodr{\'\i}guez}.} \bibinfo{year}{2025}\natexlab{}.
\newblock \showarticletitle{Mixermdm: Learnable composition of human motion diffusion models}. In \bibinfo{booktitle}{\emph{Proceedings of the Computer Vision and Pattern Recognition Conference}}. \bibinfo{pages}{12380--12390}.
\newblock


\bibitem[Shafir et~al\mbox{.}(2023)]%
        {shafir2023priormdm}
\bibfield{author}{\bibinfo{person}{Yonatan Shafir}, \bibinfo{person}{Guy Tevet}, \bibinfo{person}{Roy Kapon}, {and} \bibinfo{person}{Amit~H Bermano}.} \bibinfo{year}{2023}\natexlab{}.
\newblock \showarticletitle{Human motion diffusion as a generative prior}.
\newblock \bibinfo{journal}{\emph{arXiv preprint arXiv:2303.01418}} (\bibinfo{year}{2023}).
\newblock


\bibitem[Tevet et~al\mbox{.}(2022)]%
        {tevet2022human}
\bibfield{author}{\bibinfo{person}{Guy Tevet}, \bibinfo{person}{Sigal Raab}, \bibinfo{person}{Brian Gordon}, \bibinfo{person}{Yonatan Shafir}, \bibinfo{person}{Daniel Cohen-Or}, {and} \bibinfo{person}{Amit~H Bermano}.} \bibinfo{year}{2022}\natexlab{}.
\newblock \showarticletitle{Human motion diffusion model}.
\newblock \bibinfo{journal}{\emph{arXiv preprint arXiv:2209.14916}} (\bibinfo{year}{2022}).
\newblock


\bibitem[Wang et~al\mbox{.}(2023b)]%
        {wang2023hard}
\bibfield{author}{\bibinfo{person}{Haochen Wang}, \bibinfo{person}{Kaiyou Song}, \bibinfo{person}{Junsong Fan}, \bibinfo{person}{Yuxi Wang}, \bibinfo{person}{Jin Xie}, {and} \bibinfo{person}{Zhaoxiang Zhang}.} \bibinfo{year}{2023}\natexlab{b}.
\newblock \showarticletitle{Hard patches mining for masked image modeling}. In \bibinfo{booktitle}{\emph{Proceedings of the IEEE/CVF Conference on Computer Vision and Pattern Recognition}}. \bibinfo{pages}{10375--10385}.
\newblock


\bibitem[Wang et~al\mbox{.}(2023a)]%
        {wang2023fg}
\bibfield{author}{\bibinfo{person}{Yin Wang}, \bibinfo{person}{Zhiying Leng}, \bibinfo{person}{Frederick~WB Li}, \bibinfo{person}{Shun-Cheng Wu}, {and} \bibinfo{person}{Xiaohui Liang}.} \bibinfo{year}{2023}\natexlab{a}.
\newblock \showarticletitle{Fg-t2m: Fine-grained text-driven human motion generation via diffusion model}. In \bibinfo{booktitle}{\emph{Proceedings of the IEEE/CVF international conference on computer vision}}. \bibinfo{pages}{22035--22044}.
\newblock


\bibitem[Wang et~al\mbox{.}(2025a)]%
        {wang2025hsigpt}
\bibfield{author}{\bibinfo{person}{Yuan Wang}, \bibinfo{person}{Yali Li}, \bibinfo{person}{Xiang Li}, {and} \bibinfo{person}{Shengjin Wang}.} \bibinfo{year}{2025}\natexlab{a}.
\newblock \showarticletitle{Hsi-gpt: A general-purpose large scene-motion-language model for human scene interaction}. In \bibinfo{booktitle}{\emph{Proceedings of the Computer Vision and Pattern Recognition Conference}}. \bibinfo{pages}{7147--7157}.
\newblock


\bibitem[Wang et~al\mbox{.}(2025b)]%
        {wang2025timotion}
\bibfield{author}{\bibinfo{person}{Yabiao Wang}, \bibinfo{person}{Shuo Wang}, \bibinfo{person}{Jiangning Zhang}, \bibinfo{person}{Ke Fan}, \bibinfo{person}{Jiafu Wu}, \bibinfo{person}{Zhucun Xue}, {and} \bibinfo{person}{Yong Liu}.} \bibinfo{year}{2025}\natexlab{b}.
\newblock \showarticletitle{Timotion: Temporal and interactive framework for efficient human-human motion generation}. In \bibinfo{booktitle}{\emph{Proceedings of the Computer Vision and Pattern Recognition Conference}}. \bibinfo{pages}{7169--7178}.
\newblock


\bibitem[Wang et~al\mbox{.}(2024)]%
        {wang2024movesay}
\bibfield{author}{\bibinfo{person}{Zan Wang}, \bibinfo{person}{Yixin Chen}, \bibinfo{person}{Baoxiong Jia}, \bibinfo{person}{Puhao Li}, \bibinfo{person}{Jinlu Zhang}, \bibinfo{person}{Jingze Zhang}, \bibinfo{person}{Tengyu Liu}, \bibinfo{person}{Yixin Zhu}, \bibinfo{person}{Wei Liang}, {and} \bibinfo{person}{Siyuan Huang}.} \bibinfo{year}{2024}\natexlab{}.
\newblock \showarticletitle{Move as you say interact as you can: Language-guided human motion generation with scene affordance}. In \bibinfo{booktitle}{\emph{Proceedings of the IEEE/CVF Conference on Computer Vision and Pattern Recognition}}. \bibinfo{pages}{433--444}.
\newblock


\bibitem[Wu et~al\mbox{.}(2025)]%
        {wu2025hoi}
\bibfield{author}{\bibinfo{person}{Lin Wu}, \bibinfo{person}{Zhixiang Chen}, {and} \bibinfo{person}{Jianglin Lan}.} \bibinfo{year}{2025}\natexlab{}.
\newblock \showarticletitle{HOI-Dyn: Learning Interaction Dynamics for Human-Object Motion Diffusion}.
\newblock \bibinfo{journal}{\emph{arXiv preprint arXiv:2507.01737}} (\bibinfo{year}{2025}).
\newblock


\bibitem[Xu et~al\mbox{.}(2025a)]%
        {xu2025interact}
\bibfield{author}{\bibinfo{person}{Sirui Xu}, \bibinfo{person}{Dongting Li}, \bibinfo{person}{Yucheng Zhang}, \bibinfo{person}{Xiyan Xu}, \bibinfo{person}{Qi Long}, \bibinfo{person}{Ziyin Wang}, \bibinfo{person}{Yunzhi Lu}, \bibinfo{person}{Shuchang Dong}, \bibinfo{person}{Hezi Jiang}, \bibinfo{person}{Akshat Gupta}, {et~al\mbox{.}}} \bibinfo{year}{2025}\natexlab{a}.
\newblock \showarticletitle{Interact: Advancing large-scale versatile 3d human-object interaction generation}. In \bibinfo{booktitle}{\emph{Proceedings of the Computer Vision and Pattern Recognition Conference}}. \bibinfo{pages}{7048--7060}.
\newblock


\bibitem[Xu et~al\mbox{.}(2023)]%
        {xu2023interdiff}
\bibfield{author}{\bibinfo{person}{Sirui Xu}, \bibinfo{person}{Zhengyuan Li}, \bibinfo{person}{Yu-Xiong Wang}, {and} \bibinfo{person}{Liang-Yan Gui}.} \bibinfo{year}{2023}\natexlab{}.
\newblock \showarticletitle{Interdiff: Generating 3d human-object interactions with physics-informed diffusion}. In \bibinfo{booktitle}{\emph{Proceedings of the IEEE/CVF International Conference on Computer Vision}}. \bibinfo{pages}{14928--14940}.
\newblock


\bibitem[Xu et~al\mbox{.}(2025b)]%
        {xu2025intermimic}
\bibfield{author}{\bibinfo{person}{Sirui Xu}, \bibinfo{person}{Hung~Yu Ling}, \bibinfo{person}{Yu-Xiong Wang}, {and} \bibinfo{person}{Liang-Yan Gui}.} \bibinfo{year}{2025}\natexlab{b}.
\newblock \showarticletitle{Intermimic: Towards universal whole-body control for physics-based human-object interactions}. In \bibinfo{booktitle}{\emph{Proceedings of the Computer Vision and Pattern Recognition Conference}}. \bibinfo{pages}{12266--12277}.
\newblock


\bibitem[Xu et~al\mbox{.}(2024)]%
        {xu2024interdreamer}
\bibfield{author}{\bibinfo{person}{Sirui Xu}, \bibinfo{person}{Ziyin Wang}, \bibinfo{person}{Yu-Xiong Wang}, {and} \bibinfo{person}{Liang-Yan Gui}.} \bibinfo{year}{2024}\natexlab{}.
\newblock \showarticletitle{Interdreamer: Zero-shot text to 3d dynamic human-object interaction}.
\newblock \bibinfo{journal}{\emph{Advances in Neural Information Processing Systems}}  \bibinfo{volume}{37} (\bibinfo{year}{2024}), \bibinfo{pages}{52858--52890}.
\newblock


\bibitem[Yang et~al\mbox{.}(2024a)]%
        {yang2024codancers}
\bibfield{author}{\bibinfo{person}{Kaixing Yang}, \bibinfo{person}{Xulong Tang}, \bibinfo{person}{Ran Diao}, \bibinfo{person}{Hongyan Liu}, \bibinfo{person}{Jun He}, {and} \bibinfo{person}{Zhaoxin Fan}.} \bibinfo{year}{2024}\natexlab{a}.
\newblock \showarticletitle{CoDancers: Music-Driven Coherent Group Dance Generation with Choreographic Unit}. In \bibinfo{booktitle}{\emph{Proceedings of the 2024 International Conference on Multimedia Retrieval}}. \bibinfo{pages}{675--683}.
\newblock


\bibitem[Yang et~al\mbox{.}(2025a)]%
        {yang2025matchdance}
\bibfield{author}{\bibinfo{person}{Kaixing Yang}, \bibinfo{person}{Xulong Tang}, \bibinfo{person}{Yuxuan Hu}, \bibinfo{person}{Jiahao Yang}, \bibinfo{person}{Hongyan Liu}, \bibinfo{person}{Qinnan Zhang}, \bibinfo{person}{Jun He}, {and} \bibinfo{person}{Zhaoxin Fan}.} \bibinfo{year}{2025}\natexlab{a}.
\newblock \showarticletitle{Matchdance: Collaborative mamba-transformer architecture matching for high-quality 3d dance synthesis}.
\newblock \bibinfo{journal}{\emph{arXiv preprint arXiv:2505.14222}} (\bibinfo{year}{2025}).
\newblock


\bibitem[Yang et~al\mbox{.}(2025b)]%
        {yang2025megadance}
\bibfield{author}{\bibinfo{person}{Kaixing Yang}, \bibinfo{person}{Xulong Tang}, \bibinfo{person}{Ziqiao Peng}, \bibinfo{person}{Yuxuan Hu}, \bibinfo{person}{Jun He}, {and} \bibinfo{person}{Hongyan Liu}.} \bibinfo{year}{2025}\natexlab{b}.
\newblock \showarticletitle{Megadance: Mixture-of-experts architecture for genre-aware 3d dance generation}.
\newblock \bibinfo{journal}{\emph{arXiv preprint arXiv:2505.17543}} (\bibinfo{year}{2025}).
\newblock


\bibitem[Yang et~al\mbox{.}(2025c)]%
        {yang2025flowerdance}
\bibfield{author}{\bibinfo{person}{Kaixing Yang}, \bibinfo{person}{Xulong Tang}, \bibinfo{person}{Ziqiao Peng}, \bibinfo{person}{Xiangyue Zhang}, \bibinfo{person}{Puwei Wang}, \bibinfo{person}{Jun He}, {and} \bibinfo{person}{Hongyan Liu}.} \bibinfo{year}{2025}\natexlab{c}.
\newblock \showarticletitle{FlowerDance: MeanFlow for Efficient and Refined 3D Dance Generation}.
\newblock \bibinfo{journal}{\emph{arXiv preprint arXiv:2511.21029}} (\bibinfo{year}{2025}).
\newblock


\bibitem[Yang et~al\mbox{.}(2024b)]%
        {yang2024cohedancers}
\bibfield{author}{\bibinfo{person}{Kaixing Yang}, \bibinfo{person}{Xulong Tang}, \bibinfo{person}{Haoyu Wu}, \bibinfo{person}{Qinliang Xue}, \bibinfo{person}{Biao Qin}, \bibinfo{person}{Hongyan Liu}, {and} \bibinfo{person}{Zhaoxin Fan}.} \bibinfo{year}{2024}\natexlab{b}.
\newblock \showarticletitle{CoheDancers: Enhancing Interactive Group Dance Generation through Music-Driven Coherence Decomposition}.
\newblock \bibinfo{journal}{\emph{arXiv preprint arXiv:2412.19123}} (\bibinfo{year}{2024}).
\newblock


\bibitem[Yang et~al\mbox{.}(2024c)]%
        {yang2024beatdance}
\bibfield{author}{\bibinfo{person}{Kaixing Yang}, \bibinfo{person}{Xukun Zhou}, \bibinfo{person}{Xulong Tang}, \bibinfo{person}{Ran Diao}, \bibinfo{person}{Hongyan Liu}, \bibinfo{person}{Jun He}, {and} \bibinfo{person}{Zhaoxin Fan}.} \bibinfo{year}{2024}\natexlab{c}.
\newblock \showarticletitle{BeatDance: A Beat-Based Model-Agnostic Contrastive Learning Framework for Music-Dance Retrieval}. In \bibinfo{booktitle}{\emph{Proceedings of the 2024 International Conference on Multimedia Retrieval}}. \bibinfo{pages}{11--19}.
\newblock


\bibitem[Yang et~al\mbox{.}(2025d)]%
        {yang2025mace}
\bibfield{author}{\bibinfo{person}{Kaixing Yang}, \bibinfo{person}{Jiashu Zhu}, \bibinfo{person}{Xulong Tang}, \bibinfo{person}{Ziqiao Peng}, \bibinfo{person}{Xiangyue Zhang}, \bibinfo{person}{Puwei Wang}, \bibinfo{person}{Jiahong Wu}, \bibinfo{person}{Xiangxiang Chu}, \bibinfo{person}{Hongyan Liu}, {and} \bibinfo{person}{Jun He}.} \bibinfo{year}{2025}\natexlab{d}.
\newblock \showarticletitle{MACE-Dance: Motion-Appearance Cascaded Experts for Music-Driven Dance Video Generation}.
\newblock \bibinfo{journal}{\emph{arXiv preprint arXiv:2512.18181}} (\bibinfo{year}{2025}).
\newblock


\bibitem[Yang et~al\mbox{.}(2026)]%
        {yang2026tokendancetokentotokenmusictodancegeneration}
\bibfield{author}{\bibinfo{person}{Ziyue Yang}, \bibinfo{person}{Kaixing Yang}, {and} \bibinfo{person}{Xulong Tang}.} \bibinfo{year}{2026}\natexlab{}.
\newblock \bibinfo{title}{TokenDance: Token-to-Token Music-to-Dance Generation with Bidirectional Mamba}.
\newblock
\showeprint[arxiv]{2603.27314}~[cs.AI]
\urldef\tempurl%
\url{https://arxiv.org/abs/2603.27314}
\showURL{%
\tempurl}


\bibitem[Yi et~al\mbox{.}(2023)]%
        {yi2023generating}
\bibfield{author}{\bibinfo{person}{Hongwei Yi}, \bibinfo{person}{Hualin Liang}, \bibinfo{person}{Yifei Liu}, \bibinfo{person}{Qiong Cao}, \bibinfo{person}{Yandong Wen}, \bibinfo{person}{Timo Bolkart}, \bibinfo{person}{Dacheng Tao}, {and} \bibinfo{person}{Michael~J Black}.} \bibinfo{year}{2023}\natexlab{}.
\newblock \showarticletitle{Generating holistic 3d human motion from speech}. In \bibinfo{booktitle}{\emph{Proceedings of the IEEE/CVF Conference on Computer Vision and Pattern Recognition}}. \bibinfo{pages}{469--480}.
\newblock


\bibitem[Yuan et~al\mbox{.}(2024)]%
        {yuan2024mogents}
\bibfield{author}{\bibinfo{person}{Weihao Yuan}, \bibinfo{person}{Yisheng He}, \bibinfo{person}{Weichao Shen}, \bibinfo{person}{Yuan Dong}, \bibinfo{person}{Xiaodong Gu}, \bibinfo{person}{Zilong Dong}, \bibinfo{person}{Liefeng Bo}, {and} \bibinfo{person}{Qixing Huang}.} \bibinfo{year}{2024}\natexlab{}.
\newblock \showarticletitle{Mogents: Motion generation based on spatial-temporal joint modeling}.
\newblock \bibinfo{journal}{\emph{Advances in Neural Information Processing Systems}}  \bibinfo{volume}{37} (\bibinfo{year}{2024}), \bibinfo{pages}{130739--130763}.
\newblock


\bibitem[Zeng et~al\mbox{.}(2025)]%
        {zeng2025chainhoi}
\bibfield{author}{\bibinfo{person}{Ling-An Zeng}, \bibinfo{person}{Guohong Huang}, \bibinfo{person}{Yi-Lin Wei}, \bibinfo{person}{Shengbo Gu}, \bibinfo{person}{Yu-Ming Tang}, \bibinfo{person}{Jingke Meng}, {and} \bibinfo{person}{Wei-Shi Zheng}.} \bibinfo{year}{2025}\natexlab{}.
\newblock \showarticletitle{Chainhoi: Joint-based kinematic chain modeling for human-object interaction generation}. In \bibinfo{booktitle}{\emph{Proceedings of the IEEE/CVF Conference on Computer Vision and Pattern Recognition}}. \bibinfo{pages}{12358--12369}.
\newblock


\bibitem[Zhang et~al\mbox{.}(2025a)]%
        {zhang2025energymogen}
\bibfield{author}{\bibinfo{person}{Jianrong Zhang}, \bibinfo{person}{Hehe Fan}, {and} \bibinfo{person}{Yi Yang}.} \bibinfo{year}{2025}\natexlab{a}.
\newblock \showarticletitle{Energymogen: Compositional human motion generation with energy-based diffusion model in latent space}. In \bibinfo{booktitle}{\emph{Proceedings of the Computer Vision and Pattern Recognition Conference}}. \bibinfo{pages}{17592--17602}.
\newblock


\bibitem[Zhang et~al\mbox{.}(2024d)]%
        {zhang2024hoim3}
\bibfield{author}{\bibinfo{person}{Juze Zhang}, \bibinfo{person}{Jingyan Zhang}, \bibinfo{person}{Zining Song}, \bibinfo{person}{Zhanhe Shi}, \bibinfo{person}{Chengfeng Zhao}, \bibinfo{person}{Ye Shi}, \bibinfo{person}{Jingyi Yu}, \bibinfo{person}{Lan Xu}, {and} \bibinfo{person}{Jingya Wang}.} \bibinfo{year}{2024}\natexlab{d}.
\newblock \showarticletitle{Hoi-m\^{} 3: Capture multiple humans and objects interaction within contextual environment}. In \bibinfo{booktitle}{\emph{Proceedings of the IEEE/CVF Conference on Computer Vision and Pattern Recognition}}. \bibinfo{pages}{516--526}.
\newblock


\bibitem[Zhang et~al\mbox{.}(2023c)]%
        {zhang2023generating}
\bibfield{author}{\bibinfo{person}{Jianrong Zhang}, \bibinfo{person}{Yangsong Zhang}, \bibinfo{person}{Xiaodong Cun}, \bibinfo{person}{Yong Zhang}, \bibinfo{person}{Hongwei Zhao}, \bibinfo{person}{Hongtao Lu}, \bibinfo{person}{Xi Shen}, {and} \bibinfo{person}{Ying Shan}.} \bibinfo{year}{2023}\natexlab{c}.
\newblock \showarticletitle{Generating human motion from textual descriptions with discrete representations}. In \bibinfo{booktitle}{\emph{Proceedings of the IEEE/CVF conference on computer vision and pattern recognition}}. \bibinfo{pages}{14730--14740}.
\newblock


\bibitem[Zhang et~al\mbox{.}(2024a)]%
        {zhang2024motiondiffuse}
\bibfield{author}{\bibinfo{person}{Mingyuan Zhang}, \bibinfo{person}{Zhongang Cai}, \bibinfo{person}{Liang Pan}, \bibinfo{person}{Fangzhou Hong}, \bibinfo{person}{Xinying Guo}, \bibinfo{person}{Lei Yang}, {and} \bibinfo{person}{Ziwei Liu}.} \bibinfo{year}{2024}\natexlab{a}.
\newblock \showarticletitle{Motiondiffuse: Text-driven human motion generation with diffusion model}.
\newblock \bibinfo{journal}{\emph{IEEE transactions on pattern analysis and machine intelligence}} \bibinfo{volume}{46}, \bibinfo{number}{6} (\bibinfo{year}{2024}), \bibinfo{pages}{4115--4128}.
\newblock


\bibitem[Zhang et~al\mbox{.}(2023a)]%
        {zhang2023remodiffuse}
\bibfield{author}{\bibinfo{person}{Mingyuan Zhang}, \bibinfo{person}{Xinying Guo}, \bibinfo{person}{Liang Pan}, \bibinfo{person}{Zhongang Cai}, \bibinfo{person}{Fangzhou Hong}, \bibinfo{person}{Huirong Li}, \bibinfo{person}{Lei Yang}, {and} \bibinfo{person}{Ziwei Liu}.} \bibinfo{year}{2023}\natexlab{a}.
\newblock \showarticletitle{Remodiffuse: Retrieval-augmented motion diffusion model}. In \bibinfo{booktitle}{\emph{Proceedings of the IEEE/CVF International Conference on Computer Vision}}. \bibinfo{pages}{364--373}.
\newblock


\bibitem[Zhang et~al\mbox{.}(2023b)]%
        {zhang2024finemogen}
\bibfield{author}{\bibinfo{person}{Mingyuan Zhang}, \bibinfo{person}{Huirong Li}, \bibinfo{person}{Zhongang Cai}, \bibinfo{person}{Jiawei Ren}, \bibinfo{person}{Lei Yang}, {and} \bibinfo{person}{Ziwei Liu}.} \bibinfo{year}{2023}\natexlab{b}.
\newblock \showarticletitle{Finemogen: Fine-grained spatio-temporal motion generation and editing}.
\newblock \bibinfo{journal}{\emph{Advances in Neural Information Processing Systems}}  \bibinfo{volume}{36} (\bibinfo{year}{2023}), \bibinfo{pages}{13981--13992}.
\newblock


\bibitem[Zhang et~al\mbox{.}(2022)]%
        {zhang2022couch}
\bibfield{author}{\bibinfo{person}{Xiaohan Zhang}, \bibinfo{person}{Bharat~Lal Bhatnagar}, \bibinfo{person}{Sebastian Starke}, \bibinfo{person}{Vladimir Guzov}, {and} \bibinfo{person}{Gerard Pons-Moll}.} \bibinfo{year}{2022}\natexlab{}.
\newblock \showarticletitle{Couch: Towards controllable human-chair interactions}. In \bibinfo{booktitle}{\emph{European Conference on Computer Vision}}. Springer, \bibinfo{pages}{518--535}.
\newblock


\bibitem[Zhang et~al\mbox{.}(2025b)]%
        {zhang2025robust}
\bibfield{author}{\bibinfo{person}{Xiangyue Zhang}, \bibinfo{person}{Yifan Jia}, \bibinfo{person}{Jiaxu Zhang}, \bibinfo{person}{Yijie Yang}, {and} \bibinfo{person}{Zhigang Tu}.} \bibinfo{year}{2025}\natexlab{b}.
\newblock \showarticletitle{Robust 2D skeleton action recognition via decoupling and distilling 3D latent features}.
\newblock \bibinfo{journal}{\emph{IEEE Transactions on Circuits and Systems for Video Technology}} (\bibinfo{year}{2025}).
\newblock


\bibitem[Zhang et~al\mbox{.}(2026)]%
        {zhang2026mitigating}
\bibfield{author}{\bibinfo{person}{Xiangyue Zhang}, \bibinfo{person}{Jianfang Li}, \bibinfo{person}{Jianqiang Ren}, {and} \bibinfo{person}{Jiaxu Zhang}.} \bibinfo{year}{2026}\natexlab{}.
\newblock \showarticletitle{Mitigating Error Accumulation in Co-Speech Motion Generation via Global Rotation Diffusion and Multi-Level Constraints}. In \bibinfo{booktitle}{\emph{Proceedings of the AAAI Conference on Artificial Intelligence}}, Vol.~\bibinfo{volume}{40}. \bibinfo{pages}{12834--12842}.
\newblock


\bibitem[Zhang et~al\mbox{.}(2025d)]%
        {zhang2025semtalk}
\bibfield{author}{\bibinfo{person}{Xiangyue Zhang}, \bibinfo{person}{Jianfang Li}, \bibinfo{person}{Jiaxu Zhang}, \bibinfo{person}{Ziqiang Dang}, \bibinfo{person}{Jianqiang Ren}, \bibinfo{person}{Liefeng Bo}, {and} \bibinfo{person}{Zhigang Tu}.} \bibinfo{year}{2025}\natexlab{d}.
\newblock \showarticletitle{Semtalk: Holistic co-speech motion generation with frame-level semantic emphasis}. In \bibinfo{booktitle}{\emph{Proceedings of the IEEE/CVF International Conference on Computer Vision}}. \bibinfo{pages}{13761--13771}.
\newblock


\bibitem[Zhang et~al\mbox{.}(2025e)]%
        {zhang2025echomask}
\bibfield{author}{\bibinfo{person}{Xiangyue Zhang}, \bibinfo{person}{Jianfang Li}, \bibinfo{person}{Jiaxu Zhang}, \bibinfo{person}{Jianqiang Ren}, \bibinfo{person}{Liefeng Bo}, {and} \bibinfo{person}{Zhigang Tu}.} \bibinfo{year}{2025}\natexlab{e}.
\newblock \showarticletitle{Echomask: Speech-queried attention-based mask modeling for holistic co-speech motion generation}. In \bibinfo{booktitle}{\emph{Proceedings of the 33rd ACM International Conference on Multimedia}}. \bibinfo{pages}{10827--10836}.
\newblock


\bibitem[Zhang et~al\mbox{.}(2024b)]%
        {zhang2024kmm}
\bibfield{author}{\bibinfo{person}{Zeyu Zhang}, \bibinfo{person}{Hang Gao}, \bibinfo{person}{Akide Liu}, \bibinfo{person}{Qi Chen}, \bibinfo{person}{Feng Chen}, \bibinfo{person}{Yiran Wang}, \bibinfo{person}{Danning Li}, \bibinfo{person}{Rui Zhao}, \bibinfo{person}{Zhenming Li}, \bibinfo{person}{Zhongwen Zhou}, {et~al\mbox{.}}} \bibinfo{year}{2024}\natexlab{b}.
\newblock \showarticletitle{Kmm: Key frame mask mamba for extended motion generation}.
\newblock \bibinfo{journal}{\emph{arXiv preprint arXiv:2411.06481}} (\bibinfo{year}{2024}).
\newblock


\bibitem[Zhang et~al\mbox{.}(2025c)]%
        {zhang2025towards}
\bibfield{author}{\bibinfo{person}{Zongye Zhang}, \bibinfo{person}{Bohan Kong}, \bibinfo{person}{Qingjie Liu}, {and} \bibinfo{person}{Yunhong Wang}.} \bibinfo{year}{2025}\natexlab{c}.
\newblock \showarticletitle{Towards robust and controllable text-to-motion via masked autoregressive diffusion}. In \bibinfo{booktitle}{\emph{Proceedings of the 33rd ACM International Conference on Multimedia}}. \bibinfo{pages}{9326--9335}.
\newblock


\bibitem[Zhang et~al\mbox{.}(2024c)]%
        {zhang2024motionmamba}
\bibfield{author}{\bibinfo{person}{Zeyu Zhang}, \bibinfo{person}{Akide Liu}, \bibinfo{person}{Ian Reid}, \bibinfo{person}{Richard Hartley}, \bibinfo{person}{Bohan Zhuang}, {and} \bibinfo{person}{Hao Tang}.} \bibinfo{year}{2024}\natexlab{c}.
\newblock \showarticletitle{Motion mamba: Efficient and long sequence motion generation}. In \bibinfo{booktitle}{\emph{European Conference on Computer Vision}}. Springer, \bibinfo{pages}{265--282}.
\newblock


\bibitem[Zhao et~al\mbox{.}(2024)]%
        {zhao2025dartcontrol}
\bibfield{author}{\bibinfo{person}{Kaifeng Zhao}, \bibinfo{person}{Gen Li}, {and} \bibinfo{person}{Siyu Tang}.} \bibinfo{year}{2024}\natexlab{}.
\newblock \showarticletitle{DartControl: A diffusion-based autoregressive motion model for real-time text-driven motion control}.
\newblock \bibinfo{journal}{\emph{arXiv preprint arXiv:2410.05260}} (\bibinfo{year}{2024}).
\newblock


\bibitem[Zhi et~al\mbox{.}(2023)]%
        {zhi2023livelyspeaker}
\bibfield{author}{\bibinfo{person}{Yihao Zhi}, \bibinfo{person}{Xiaodong Cun}, \bibinfo{person}{Xuelin Chen}, \bibinfo{person}{Xi Shen}, \bibinfo{person}{Wen Guo}, \bibinfo{person}{Shaoli Huang}, {and} \bibinfo{person}{Shenghua Gao}.} \bibinfo{year}{2023}\natexlab{}.
\newblock \showarticletitle{Livelyspeaker: Towards semantic-aware co-speech gesture generation}. In \bibinfo{booktitle}{\emph{Proceedings of the IEEE/CVF international conference on computer vision}}. \bibinfo{pages}{20807--20817}.
\newblock


\bibitem[Zhou et~al\mbox{.}(2024)]%
        {zhou2024emdm}
\bibfield{author}{\bibinfo{person}{Wenyang Zhou}, \bibinfo{person}{Zhiyang Dou}, \bibinfo{person}{Zeyu Cao}, \bibinfo{person}{Zhouyingcheng Liao}, \bibinfo{person}{Jingbo Wang}, \bibinfo{person}{Wenjia Wang}, \bibinfo{person}{Yuan Liu}, \bibinfo{person}{Taku Komura}, \bibinfo{person}{Wenping Wang}, {and} \bibinfo{person}{Lingjie Liu}.} \bibinfo{year}{2024}\natexlab{}.
\newblock \showarticletitle{Emdm: Efficient motion diffusion model for fast and high-quality motion generation}. In \bibinfo{booktitle}{\emph{European Conference on Computer Vision}}. Springer, \bibinfo{pages}{18--38}.
\newblock


\bibitem[Zou et~al\mbox{.}(2024)]%
        {zou2024parco}
\bibfield{author}{\bibinfo{person}{Qiran Zou}, \bibinfo{person}{Shangyuan Yuan}, \bibinfo{person}{Shian Du}, \bibinfo{person}{Yu Wang}, \bibinfo{person}{Chang Liu}, \bibinfo{person}{Yi Xu}, \bibinfo{person}{Jie Chen}, {and} \bibinfo{person}{Xiangyang Ji}.} \bibinfo{year}{2024}\natexlab{}.
\newblock \showarticletitle{Parco: Part-coordinating text-to-motion synthesis}. In \bibinfo{booktitle}{\emph{European Conference on Computer Vision}}. Springer, \bibinfo{pages}{126--143}.
\newblock


\end{thebibliography}

\end{document}